\documentclass{article}  % 使用文章类模板

\usepackage{PRIMEarxiv}  % 引入 PRIMarxiv 样式包
\usepackage[utf8]{inputenc} % 允许使用 UTF-8 编码输入
\usepackage[T1]{fontenc}    % 使用 8 位 T1 字体
\usepackage{hyperref}       % 支持超链接功能
\usepackage{url}            % 简单的 URL 排版
\usepackage{booktabs}       % 提供专业质量的表格排版
\usepackage{amsfonts}       % 提供黑板数学符号（如集合符号等）
\usepackage{nicefrac}       % 提供简洁的分数符号（如 1/2 等）
\usepackage{microtype}      % 微排版优化
\usepackage{lipsum}         % 生成随机文本
\usepackage{fancyhdr}       % 用于定制页眉页脚
\usepackage{graphicx}       % 插入图形的包
\usepackage{amsmath}
\usepackage{float}

\graphicspath{{media/}}     % 图像存放在 media/ 文件夹下

\title{CarbonChat: Large Language Model-Based Corporate Carbon Emission Analysis and Climate Knowledge Q\&A System
}

\author{
Zhixuan Cao \\
  College of Future Information Technology\\
  Shijiazhuang University\\
  Shijiazhuang, Hebei Province 050035 \\
  \texttt{20221308007@sjzc.edu.cn} \\
    \And
Ming Han \\
  College of Future Information Technology\\
  Shijiazhuang University\\
  Shijiazhuang, Hebei Province 050035 \\
  \texttt{han\_ming2008@126.com} \\
    \And
Jingtao Wang \\
  College of Future Information Technology\\
  Shijiazhuang University\\
  Shijiazhuang, Hebei Province 050035 \\
  \texttt{wjt326@163.com} \\
    \And
Meng Jia \\
  College of Future Information Technology\\
  Shijiazhuang University\\
  Shijiazhuang, Hebei Province 050035 \\
  \texttt{18811727137@163.com} \\
}

\begin{document}

\maketitle  % 生成标题

\begin{abstract}
As the impact of global climate change intensifies, corporate carbon emissions have become a focal point of global attention. In response to issues such as the lag in climate change knowledge updates within large language models, the lack of specialization and accuracy in traditional augmented generation architectures for complex problems, and the high cost and time consumption of sustainability report analysis, this paper proposes CarbonChat: Large Language Model-based corporate carbon emission analysis and climate knowledge Q\&A system, aimed at achieving precise carbon emission analysis and policy understanding.First, a diversified index module construction method is proposed to handle the segmentation of rule-based and long-text documents, as well as the extraction of structured data, thereby optimizing the parsing of key information.Second, an enhanced self-prompt retrieval-augmented generation architecture is designed, integrating intent recognition, structured reasoning chains, hybrid retrieval, and Text2SQL, improving the efficiency of semantic understanding and query conversion.Next, based on the greenhouse gas accounting framework, 14 dimensions are established for carbon emission analysis, enabling report summarization, relevance evaluation, and customized responses.Finally, through a multi-layer chunking mechanism, timestamps, and hallucination detection features, the accuracy and verifiability of the analysis results are ensured, reducing hallucination rates and enhancing the precision of the responses.
\end{abstract}

% 关键词部分可删除

\section*{Introduction}

Global climate change is a widespread and profound systemic transformation of the economic and social systems, involving various sectors such as energy structure, industrial processes, transportation, as well as ecosystem carbon sinks, engineered carbon sinks, and more. In the context of current global climate change, sustainability issues have gradually evolved into a critical topic. Enterprises are facing management challenges in measuring, reporting, and verifying greenhouse gas emissions during their operations. The Greenhouse Gas  Protocol(GHG Protocol) provides methods and standards for greenhouse gas accounting, enabling an accurate and fair reflection of an enterprise's carbon emissions.
However, the lengthy corporate carbon emission reports and government policies related to carbon emissions (often exceeding 30 pages) pose challenges for enterprises, investors, and the general public in terms of understanding and analysis. At the same time, reliance on third-party rating agencies is not always effective, as the services provided by these agencies are often costly, lack transparency, and use differing standards when evaluating corporate carbon emission reports. This further increases the complexity and uncertainty of the evaluation process.\par
With the development of large language model (LLM) technology, deep learning models based on massive text data have demonstrated powerful language processing and generation capabilities. In 2022, OpenAI's ChatGPT\cite{Achiam2023GPT4TR}, based on the GPT-3.5 architecture, made groundbreaking progress in general knowledge question answering. Subsequently, GPT-4 and its multimodal version, GPT-4V, further enhanced the model's capabilities. Meanwhile, Alibaba's Qwen\cite{Bai2023QwenTR}has also excelled in Chinese tasks, achieving performance comparable to the internationally leading GPT-4 and GPT-3 models. Despite significant advancements in the research field, large language models often struggle to provide precise answers when addressing domain-specific issues, due to a lack of domain knowledge or up-to-date data. Therefore, building an intelligent system based on large language models can more efficiently conduct data analysis and professional question answering, bridging the gaps in traditional question-answering systems.

\section*{Results}

This paper proposes Large Language Model-based enterprise carbon emission analysis system and global climate change knowledge Q\&A system, as shown in Figures 1 and 2. To address the issues of insufficient data referencing and integration of policies and regulations in carbon emission-related knowledge and regulatory searches, this paper proposes a diversified indexing module construction method. Through this method, a high-quality carbon emission knowledge and regulatory search dataset is built, significantly improving the accuracy and reliability of knowledge retrieval, and providing users with more comprehensive and accurate policy and regulatory references.\par
Additionally, this paper proposes a Self-Prompting Retrieval-Enhanced Generation (Self-Prompting RAG) architecture to improve the quality and authenticity of responses from large language models through retrieval and self-prompting. The architecture includes four modules: intent recognition (question classification), Structured chain-of-thought (COT) prompting, vector database hybrid retrieval, and Text2SQL system. By incorporating prompt engineering techniques, high-quality prompts are constructed to ensure seamless coordination among modules, enabling efficient semantic understanding and query transformation. The system can rapidly handle multi-dimensional complex issues while effectively enhancing the traceability and relevance of answers to the questions, reducing hallucination rates during the generation process, and further improving the overall performance of the system.\par
The enterprise carbon emission analysis system is based on the widely used international Greenhouse Gas Protocol (GHG Protocol) \cite{ghgprotocol}, and integrates modules for report summarization, relevance assessment, and customized question answering. The system conducts an in-depth analysis of enterprise sustainability reports across fourteen dimensions, including energy usage analysis, accurately quantifying their greenhouse gas emissions and generating structured and traceable analysis reports. It also combines carbon emission target setting and risk management to provide carbon emission assessments and predictions. The system automatically calculates the alignment of reports with the GHG Protocol and generates a compliance evaluation report, offering key references for enterprises to understand their report content and improve standardization.
Through diversified indexing module construction, modality tagging, and hallucination detection features, the system significantly reduces the risk of hallucination generation, ensuring the accuracy and verifiability of results. Additionally, the system supports personalized question parsing and semantic search, allowing it to flexibly address complex scenarios and provide comprehensive support for enterprises to optimize their carbon emission analysis strategies.\par
\begin{figure}[h]  
    \centering  % 使图片居中
    \includegraphics[width=0.9\textwidth]{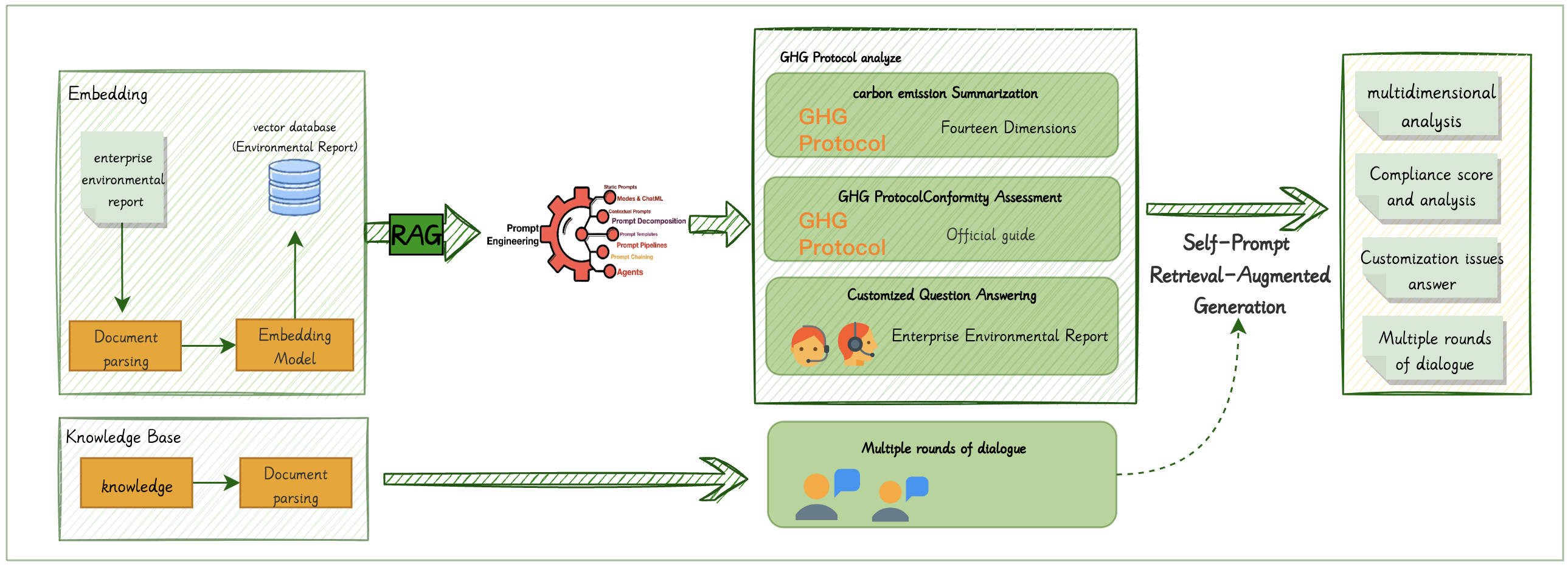}  % 插入图片并设置宽度
    \caption{Architecture Diagram of the Corporate Carbon Emission Analysis System}  % 图片标题
    \label{fig:arch1}  % 图片标签，用于引用
\end{figure}
In summary, the main contributions of this paper include:
\begin{itemize}
\item The paper proposes a diversified index module construction method and establishes a search dataset for carbon emissions-related knowledge and regulations. This approach addresses the issue of insufficient data in terms of referencing and integrating policy and regulatory knowledge, thereby enhancing the retrieval quality of carbon emissions-related knowledge.y enhancing the retrieval quality of carbon emission-related information.
\item The paper proposes a Self-Prompting Retrieval-Augmented Generation architecture, which integrates components such as intent recognition (question classification), question parsing, hybrid retrieval from vector databases, structured Chain-of-Thought (COT), and a Text2SQL system. This architecture achieves efficient semantic understanding and query transformation, enhancing the ability to handle complex queries and reducing hallucination rates in large models.
\item This paper proposes a Text2SQL system that utilizes Large Language Model to parse user queries. It generates SQL through step-by-step reasoning guided by Chain-of-Thought (COT) examples retrieved via Retrieval-Augmented Generation (RAG). The system combines semantic information with the database schema for pattern matching and semantic mapping, thereby generating structured SQL queries. Security validation is introduced to restrict high-risk operations, while automatic SQL repair and structural rewriting are employed to optimize the queries, ensuring both their safety and accuracy.
\item This paper develops a structured prompt engineering method based on the most widely used greenhouse gas accounting framework internationally, the GHG Protocol. The proposed method can automatically analyze corporate carbon emission reports across various dimensions, quantify their compliance with the international greenhouse gas accounting system, and implement benchmark testing and automatic measurement standards for disclosure quality.
\end{itemize}

\begin{figure}[ht] 
    \centering  % 使图片居中
    \includegraphics[width=0.5\textwidth]{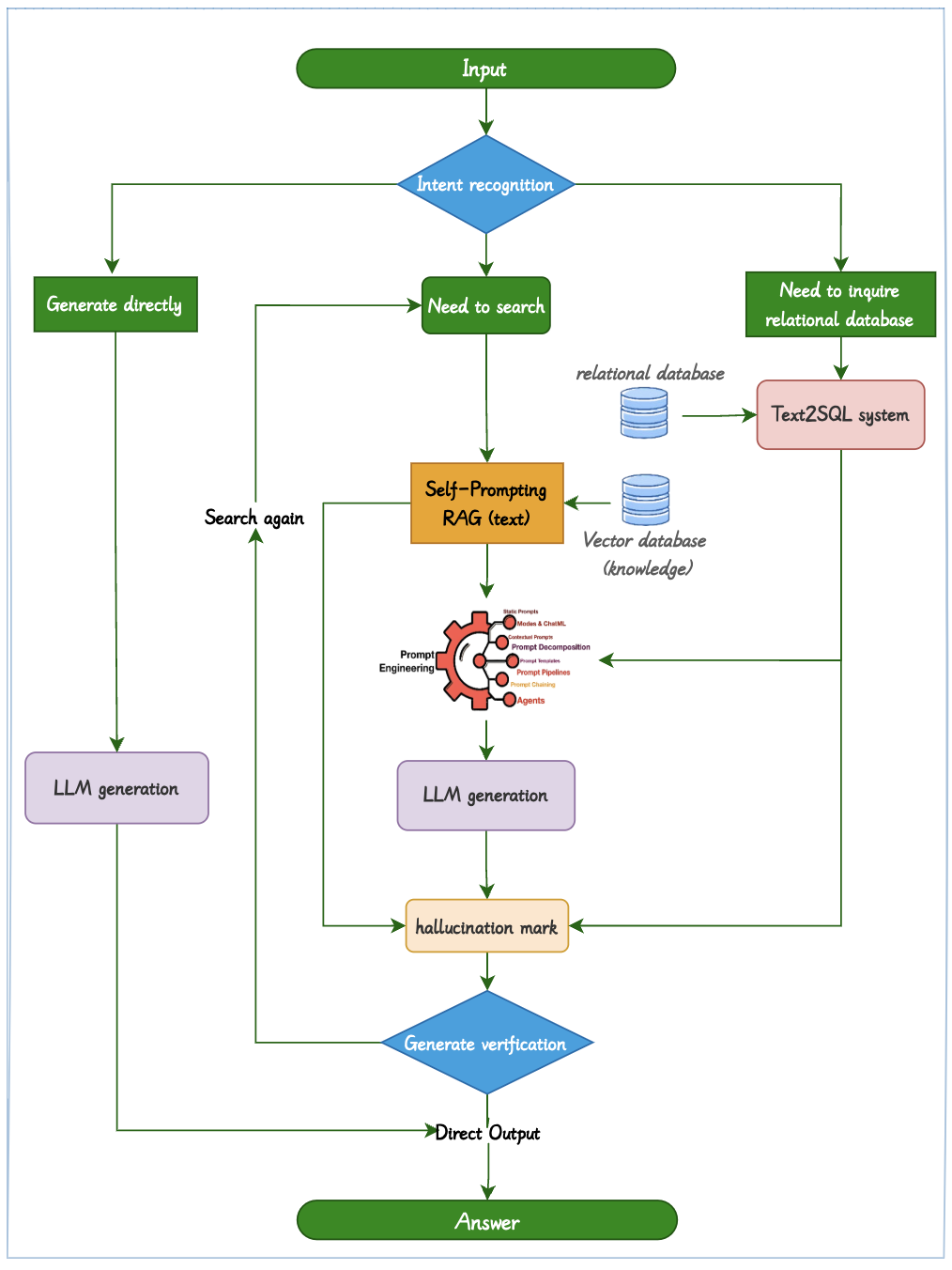}  % 插入图片并设置宽度
    \caption{Architecture Diagram of the Global Climate Change Knowledge Q\&A System}  % 图片标题
    \label{fig:arch2}  % 图片标签，用于引用
\end{figure}

\section*{Related work}

Recently, Vaghefi et al. (2023) introduced the CHATCLIMATE chatbot based on the IPCC report \cite{Vaghefi2023chatClimateGC}. This system integrates natural language processing technology and the latest scientific reports to provide accurate and timely information, promoting climate change research, policy decision-making, and public engagement. Jingwei Ni et al. (2023) developed an automated analysis system for corporate sustainability reports based on the TCFD framework\cite{Ni2023CHATREPORTDS}, aimed at improving analysis efficiency, transparency, and accuracy. Haiqi Jiang et al. proposes a framework for predicting the EU and China carbon markets\cite{Jiang2024CarbonPF}, combining time series models and large language models to optimize predictions, helping companies and governments with effective planning. Ahmad Faiz et al. designed an end-to-end carbon footprint prediction model\cite{Faiz2023LLMCarbonMT}, analyzing the carbon emissions of large language models. Xiaowei Xiang et al. launched a large model-driven science and technology policy regulation Q\&A system, enhancing accuracy and efficiency in answering queries\cite{KXTS202409007}. Yanping Zhang et al. proposes a multi-strategy retrieval-augmented generation (MSRAG) method \cite{JSJY20240828002}, optimizing military knowledge Q\&A, though improvements are still needed in handling long texts and query rewriting strategies. Yijin Wei et al. developed an agricultural policy Q\&A system based on ChatGLM2-6B and Langchain \cite{KYXH202404011} but further improvements are required in handling complex associative issues and adapting to out-of-domain problems.\par
The relevance of responses and hallucinations in large language model remain key limitations \cite{Ji2022SurveyOH}. Related research has proposes some preliminary solutions: 1) SelfCheckGPT \cite{Manakul2023SelfCheckGPTZB}, which utilizes the LLM's own capabilities to sample multiple responses and identify fabricated statements by measuring the consistency of information between them. 2) Construction of inference classifiers: such as the Fact Inconsistency Benchmark (FIB) \cite{Tam2022EvaluatingTF}, the ExHalder framework for detecting hallucinations in news headlines \cite{Shen2023WhyIT}, the HaluEval large-scale hallucination evaluation benchmark\cite{Li2023HaluEvalAL}, and fact-checking for AI-generated reports \cite{Mahmood2023FactCheckingOA}. 3) Uncertainty measurement: evaluating generated text using text generation models like BARTScore \cite{Yuan2021BARTScoreEG}, and exploring the uncertainty of known unknowns with large language models through KoK \cite{Amayuelas2023KnowledgeOK}. These efforts indicate potential methods to mitigate hallucinations in LLMs, but the issue remains an open research question.

\section*{Methods}
\subsection*{Global Climate Change Knowledge Q\&A System}
\subsubsection*{Diversified Index Module Construction}
During the document segmentation process, larger chunks provide more context, enhancing comprehension but increasing processing time. Smaller chunks improve retrieval recall and reduce time, but may lack sufficient context.\par 
This system adopts different segmentation strategies for different types of documents, with the overall process illustrated in Figure 3.\par 
1.Document Tree-Based Chunking\par
For well-structured documents (e.g., corporate reports, academic papers), the document is hierarchically divided according to its structure, such as by sentences, paragraphs, and headings, forming multiple levels. Key information from articles and paragraphs is extracted, and retrieval is performed sequentially at each level. An explicit stack iteration is used to replace recursion, allowing the system to search for node titles level by level. This addresses potential stack overflow issues caused by deep recursion and optimizes path compression. By updating the parent node information along the entire path in a single computation, the system shortens the path length and reduces redundant computations of title paths, as illustrated in Algorithm 1.\par

\begin{table}[ht]
\centering
\resizebox{\textwidth}{!}{  % 将表格缩放到页面宽度
\begin{tabular}{|l|}
\hline
\textbf{Algorithm 1: Document Tree (DT)} \\
\hline
1. Initialize Paragraph Collection: \( S \leftarrow \emptyset \) \\
2. Let: Node be \( n_i \), if it is a title \( n_i = f_{\text{title}} \) \\
$f_{\text{title}} = 
\begin{cases} 
1, & \text{if } \text{para\_type}[:5] = \text{"title"} \\
0, & \text{otherwise}
\end{cases}$ \\
3. Path Compression Optimization \\
For a node \( n_i \), its root node path is defined as: \\
$\text{root}(n_i) =
\begin{cases}
n_i, & \text{if } \text{parent}(n_i) = \text{None} \\
\text{root}(\text{parent}(n_i)), & \text{otherwise}
\end{cases}$\\
Update Path: \\
$\text{parent}(n_i) \gets \text{root}(n_i), \quad \forall n_i \in P(n_i)$ \\
The title path after path compression optimization is: \\
$P(n_i) =
\begin{cases}
[\text{text}(n_i)], & \text{if } \text{root}(n_i) = n_i \\
T(\text{root}(n_i)) + [\text{text}(n_i)], & \text{otherwise}
\end{cases}$ \\
4. Iteration \\
Let: \( P(i) \) be the title hierarchy of node \( n_i \) (a list of title paths from the root node to \( n_i \)). \\
\( \text{parent}(i) \) is the parent node of \( n_i \), and \( \text{text}(i) \) is the title text of node \( n_i \). \\
$\text{Initialize: } S \gets [i], \quad P(i) \gets []$ \\
$\text{While } S \neq []:
\begin{aligned}
    & j \gets S.\text{pop}() \\
    & \text{If } \text{parent}(j) = \text{None}: \\
    & \quad S.\text{append}(\text{parent}(j)) \\
    & P(j) \gets T(\text{parent}(j)) + [\text{text}(j)]
\end{aligned}$ \\
Initialize a stack \( S \) and push the target node \( i \) onto the stack. For each iteration, pop a node \( j \) from the stack. \\If \( j \) has a parent node, push the parent node onto the stack. Update \( P(j) \) to be the title hierarchy of its parent node, followed by the title of node \( j \).\\
5. Merging \\
$f_{\text{merge}} = 
\begin{cases} 
\text{True}, & \text{if } C_{\text{ni}} = 1 \text{ and } (f_{\text{title}} = 1 \text{ or current node is last}) \\
\text{False}, & \text{otherwise}
\end{cases}$ \\
If the condition is True, for each node \( n_i \), collect its content (if \( n_i \) is a body node). \\
This content is collected by traversing all body nodes \( n_j \) under \( n_i \),\\ performing a union operation on their content \( C(n_j) \), and aggregating them as the body content of the current paragraph. \\
$C(n_i) = \bigcup_{j \in \text{Content}(n_i)} C(n_j)$ \\
6. The Paragraph Collection \( S \) includes the paths and body content of all nodes in the document, in the form of: \\
$S = \bigcup_{i=1}^N \left( P(n_i), \bigcup_{j \in \text{Content}(n_i)} C(n_j) \right)$ \\
7. Convert the content of each node \( C(n_i) \) into a vector \( V(n_i) \): \\
$V(n_i) = \text{BGE-m3-Embedding}(C(n_i))$ \\
\hline
\end{tabular}}
\caption{\label{tab:example1}Algorithm 1: Document Tree(DT).}
\end{table}

2.Rule-Based Chunking and Semantic-Based Chunking\par
For documents with fixed formats, such as regulatory policy documents, financial reports, and bidding documents, rule-based chunking is more suitable. These documents often contain fixed clauses or section numbers (e.g., "Contract Clauses", "Appendix") and itemized numbers (e.g., "Article 1", "Article 2"), which can be precisely located using predefined rules.\par 
However, for documents that are more complex in content and lack a fixed structure, such as global climate change policy analysis reports and news sentiment reports, which are primarily in natural language and lack clear section divisions, semantic chunking is a more appropriate approach. In this paper, semantic chunking is based on the SeqModel proposes by Zhang et al. \cite{Zhang2021SequenceMW}. The model first uses BERT to jointly encode multiple consecutive sentences, directly modeling the dependencies between sentences to capture contextual semantic representations. Then, it predicts the segmentation boundaries of each sentence to accurately identify topic shifts. Finally, the model integrates adaptive sliding window techniques, dynamically adjusting the window size based on the content of the text, thereby improving inference efficiency while ensuring chunking accuracy. Semantic chunking effectively uncovers implicit semantics and integrates cross-paragraph information, making the chunking results better aligned with the logical structure of the topic content. \par
3.Paragraph-Based Chunking and Sliding Chunking for Long Text Blocks\par
Paragraph-based chunking is a document-structure-based segmentation approach, which divides the text by identifying natural paragraph boundaries within the document. This method is suitable for long texts that have clear paragraph divisions, such as white papers or technical documentation.
If a paragraph is too lengthy to be directly used for subsequent processing, sliding window chunking is employed for further refinement. Sliding window chunking ensures that each chunk contains a certain amount of overlapping content, thereby preserving the contextual continuity between paragraphs. Meanwhile, Large Language Model is utilized to generate high-quality summaries for each chunk, extracting the core keywords from the long text. The generated summaries are then converted into vector representations for semantic search.
Through multi-level chunking and semantic processing mechanisms, this approach addresses the issue of information omission in long texts, ensuring both the completeness and accuracy of the content in subsequent retrieval and generation tasks.\par 
4.Processing of Structured Data\par
In document recognition and extraction, for the processing of structured data (such as images, formulas, and tables), accurate data extraction is achieved through classification methods. For images, key content is extracted using a multimodal annotation approach \cite{Bai2023QwenVLAF}. The extracted text is then transformed into semantic vectors using embedding models and stored in a vector database. For formulas, the Mathpix tool is used to parse them into LaTeX expressions \cite{mathpix}, and the semantic information of the formulas is then extracted using large language models, converting them into vectors for storage, facilitating subsequent semantic retrieval and analysis.
For table data processing, the main challenges include page-break disconnections, lack of headers, and merged cells. To address the issue of tables spanning multiple pages, table boundaries and their context (e.g., footer and header text) are detected to determine table continuity, and document layout analysis (such as PDF layout or docx XML attributes) is used to track table numbering and titles to establish page-to-page connections. For tables without headers, a content-based classification model is used to predict the header position, and contextual information is employed to automatically complete the header fields. The extraction of merged cells depends on the parsing of PDF XML attributes, particularly by reading the gridSpan and vMerge attributes to determine cell merging in the horizontal and vertical directions, ensuring accurate data reclassification.
To ensure the accuracy of table extraction, the system combines XML attributes and text extraction information for verification and fusion. The precise row and column data of the table are obtained by analyzing the XML structure of docx files. In case of data conflicts, content similarity matching and field validation logic are used to select high-confidence results. The processed table content is finally stored in the MySQL database, while the semantic information of images and formulas is stored in the vector database to support efficient semantic retrieval.
\begin{figure}[ht]  % 强制图片尽量放置在当前页面
    \centering  % 使图片居中
    \includegraphics[width=0.8\textwidth]{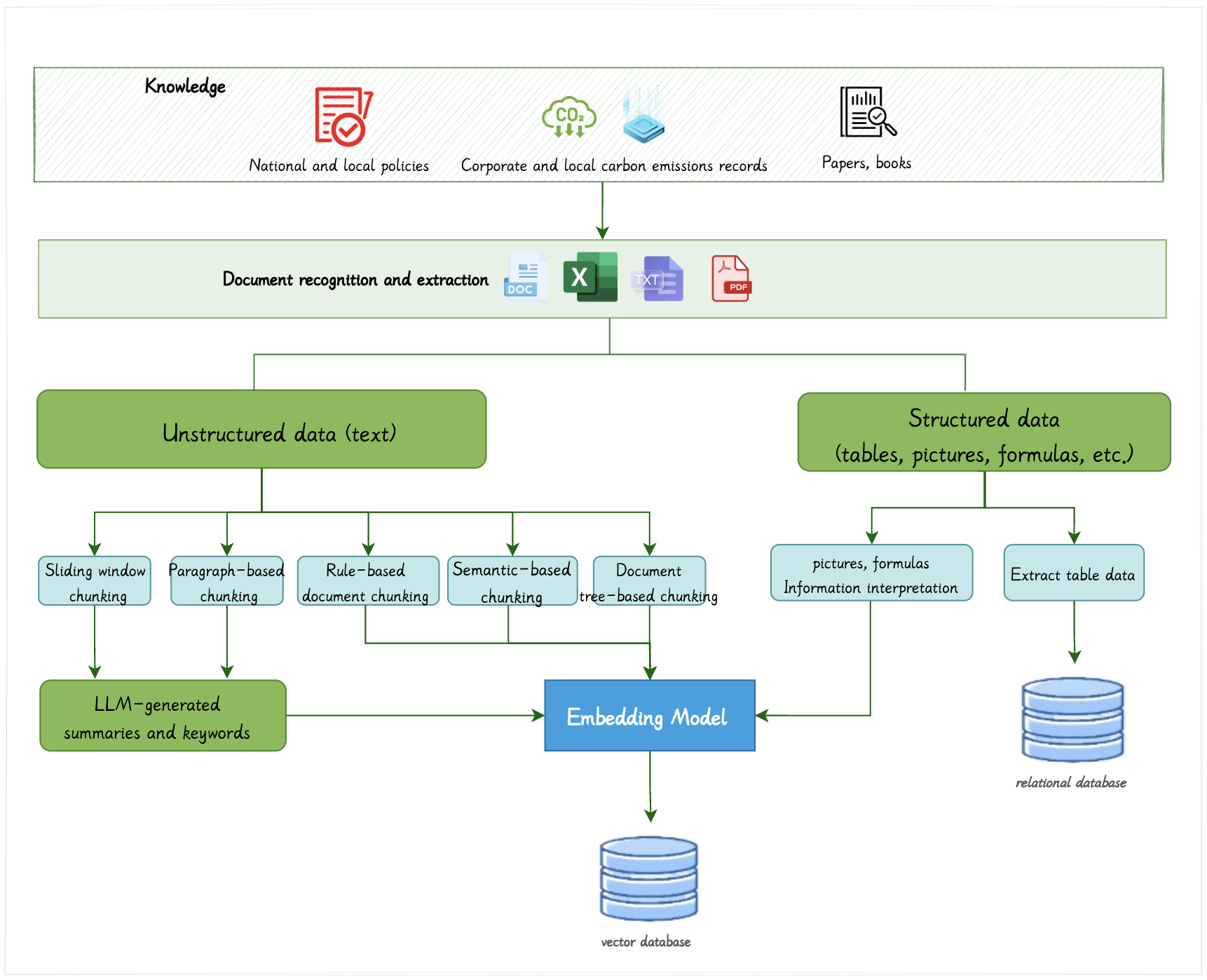}  % 插入图片并设置宽度
    \caption{Document Parsing Module}  % 图片标题
    \label{fig:en_image/4.png}  % 图片标签，用于引用
\end{figure}

\subsubsection*{Self-Prompting Enhanced Retrieval-Generation Architecture}
Traditional retrieval methods still face challenges in accuracy and authority when addressing knowledge-based questions in the field of carbon emissions. To address this issue, this paper proposes a Self-Prompting Enhanced Retrieval-Generation Architecture, which includes intent recognition, structured chain-of-thought (COT) reasoning (including query rewriting, diverse expressions, few-shot COT prompts + pre-answering, key sentence extraction, and reordering), hybrid retrieval, and the Text2SQL system. Combined with high-quality prompt engineering, this architecture achieves efficient semantic understanding and query transformation. The text-based retrieval process is illustrated in Figure 4.The prompt for this module can be found in Appendix 1.\par
By rewriting user queries, we first decompose them into simpler, more answerable subqueries. Then, we identify and eliminate ambiguities in complex queries, recognize and extract the basic intent and core conceptual elements of the query, and create a high-level representation that captures the essential meaning while removing specific details. The rewritten queries expand the semantic space, helping to uncover more potentially relevant information, especially when user queries are vague or unclear. This method compensates for gaps in user input.\par
By generating multiple expressions, the system handles different wordings and phrasings in user queries, thereby enhancing the recall and precision of the retrieval system.\par
Few-shot COT prompts assist the model in better simulating human reasoning processes, generating more accurate and contextually coherent preliminary answers. By simulating the reasoning chain, the model relies not only on semantic matching but also on knowledge and logic for inference, guiding it to gradually deduce conclusions. This approach reduces the likelihood of generating incorrect answers while helping the model better understand complex problems.\par
By extracting key sentences from the preliminary answers, more precise core information can be distilled. This enables the system to match each of the user's key points, improving the relevance of the retrieval. Key sentences preserve the complete semantics of the original text, whereas keywords, typically individual terms, may capture part of the core information but risk losing context, leading to partial or inaccurate understanding of the retrieved information.\par
BM25 is a classical retrieval model\cite{robertson1995okapi} based on Term Frequency-Inverse Document Frequency (TF-IDF), which effectively handles word matching between documents and queries, and is particularly efficient in processing short-text queries, providing highly accurate retrieval results. Embedding, on the other hand, captures deeper semantic similarities, compensating for BM25's limitation in semantic matching. Through similarity calculations based on word or sentence embeddings, the Embedding model can identify documents that are semantically closer. By combining BM25 and Embedding in a hybrid retrieval approach, the system integrates the advantages of both term frequency matching and semantic matching. This hybrid model ensures high retrieval accuracy while enhancing semantic coverage.\par
Finally, the system employs a re-ranking model, BGE-reranker-large (BAAI General Embeddings) \cite{Xiao2023CPackPR}, to concatenate and reorder the retrieval results, ensuring that the most relevant results are returned to the user. The overall process is shown in Figure 4.\par
Use BM25 and BGE M3-Embedding model \cite{Chen2024BGEMM}to perform hybrid retrieval, and select Top5.\\
$$\mathrm{Score}(q,d)=\lambda\cdot\frac1{BM25(q,d)+c}+(1-\lambda)\cdot\frac1{Embedding(q,d)+c}\    (1)$$   \par
Here: \( \lambda \in [0,1] \) is the weight of the BM25 model, and \( 1-\lambda \) is the weight of the BGE M3-Embedding model. BM25(q,d) and Embedding(q,d) are the rankings of document d in the BM25 and BGE M3-Embedding models. The lower the ranking (i.e. the larger the number), the less relevant the document is. c is used to prevent the denominator from being zero when the ranking is 0, and to control the smoothness of the ranking. The usual value is 60. The larger the c value, the smaller the impact of low-ranking documents.  \par
After the intelligent retrieval results are spliced, the bge-reranker-large rearrangement model is used to rearrange, and the Top5 are selected.
$$\mathrm{R_{final}=BGEreranker(TopN(\bigcup_{q_i\in rewrite(Q)}score(q_i,d),\bigcup_{k_j\in key_sentences(COT(Q))}score(k_j,d)))}\    (2)$$

Here:  rewrite(Q): This refers to rewriting the original query $Q$ to generate multiple query variants $q_i$, thereby expanding the search scope and increasing the diversity of the retrieval.
score($q_i,d$): This refers to the score between each query variant $q_i$ and the document $d$. The score can be based on some similarity measure, such as BM25 or embedding-based matching.key\_sentences(COT(Q)): This refers to the key sentences $k_j$ within the reasoning chain (COT) generated based on the query $Q$. The reasoning chain may consist of a series of logical steps, and extracting key sentences from this chain enhances the accuracy and relevance of the reasoning.
score($k_j,d$): This computes the score between each key sentence $k_j$ and the document $d$. Similar to the query variant scores, it measures the degree of match between the sentence and the document.
\begin{figure}[ht]  % 强制图片尽量放置在当前页面
    \centering  % 使图片居中
    \includegraphics[width=0.85\textwidth]{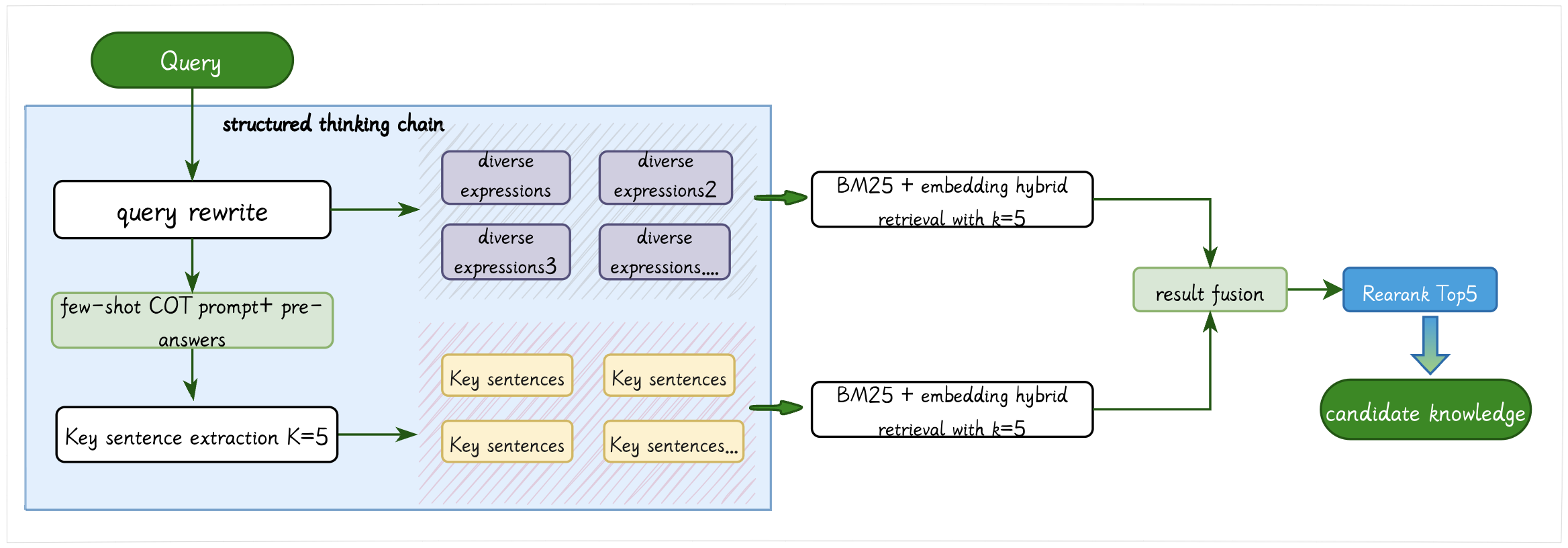}  % 插入图片并设置宽度
    \caption{Self-Prompting Enhanced Retrieval-Augmented Generation Architecture(Text Content)}  % 图片标题
    \label{fig:en_image/3.png}  % 图片标签，用于引用
\end{figure}

\subsubsection*{Text2SQL System}
The corporate carbon emissions knowledge query system requires real-time querying and analysis of large volumes of data, efficiently integrating with enterprise databases. Text2SQL, as a significant direction in natural language processing (NLP), is a technique that converts user queries in natural language into Structured Query Language (SQL). The core of this technology lies in the fusion of natural language understanding with database syntax structure, which helps lower the technical barrier for users interacting with databases. However, traditional Text2SQL methods still face the following challenges in practical applications:1) Complex queries, such as multi-table queries and queries across multiple time periods, are difficult to parse;2) The completeness and accuracy of the generated query results are not always high;3) The lack of syntax validation and security checks often leads to execution errors or security risks.To address these issues, this paper reconstructs an optimized Text2SQL system, combining the strong semantic understanding capabilities of large language model with schema awareness of databases to overcome the limitations of traditional methods.\par
The system architecture is shown in Figure 5, consisting of the following key steps:\par
Step 1: User Query Parsing and Information Extraction\par
First, the large language model is used to parse the user query, performing concurrent tasks of time extraction and data table identification. The system then retrieves similar question examples using RAG to guide the model in step-by-step reasoning to generate the SQL query. Each example demonstrates the reasoning chain from the question to the SQL query, helping the model learn how to generate accurate SQL statements without a large amount of annotated data.\par
Step 2: Generate Preliminary SQL Query by Integrating Database Schema (Table Definitions)\par
The system combines the extracted semantic information with the database schema structure to generate a preliminary SQL query. In this step, schema matching and semantic mapping are performed to ensure that the generated SQL query is consistent with the logical structure of the database. By analyzing the relationships between database tables and field constraints, the system can select appropriate tables and fields based on the query semantics, avoiding mismatches between the semantics and the table structure.\par
Step 3: SQL Query Security Validation and Syntax Optimization\par
After the preliminary SQL query is generated, the system introduces two important validation steps:\par
Security Validation: By setting up whitelists or rule-based filters, high-risk operations (such as DELETE, INSERT, UPDATE) are restricted to minimize their impact on the database, thereby enhancing the system’s security. This step uses static analysis to parse and check the SQL query's syntax tree.\par
Syntax Optimization: The system performs syntax validation on the preliminary SQL query to detect potential errors, such as missing join conditions or incorrect logical operators. Syntax optimization utilizes SQL correction and structured rewriting methods \cite{Pourreza2023DINSQLDI} to improve the completeness and accuracy of the generated query.\par
Step 4: Query Execution and Result Return\par
After the optimized SQL query is validated, the system performs a final syntax check, executes the query, and returns the result. The system integrates the semantic information parsed by the large language model, the database schema, and the execution logic of the SQL query at multiple levels to ensure both the efficiency and accuracy of the query. An example of a query is shown in Figure 6.\par
\begin{figure}[ht]  % 强制图片尽量放置在当前页面
    \centering  % 使图片居中
    \includegraphics[width=0.8\textwidth]{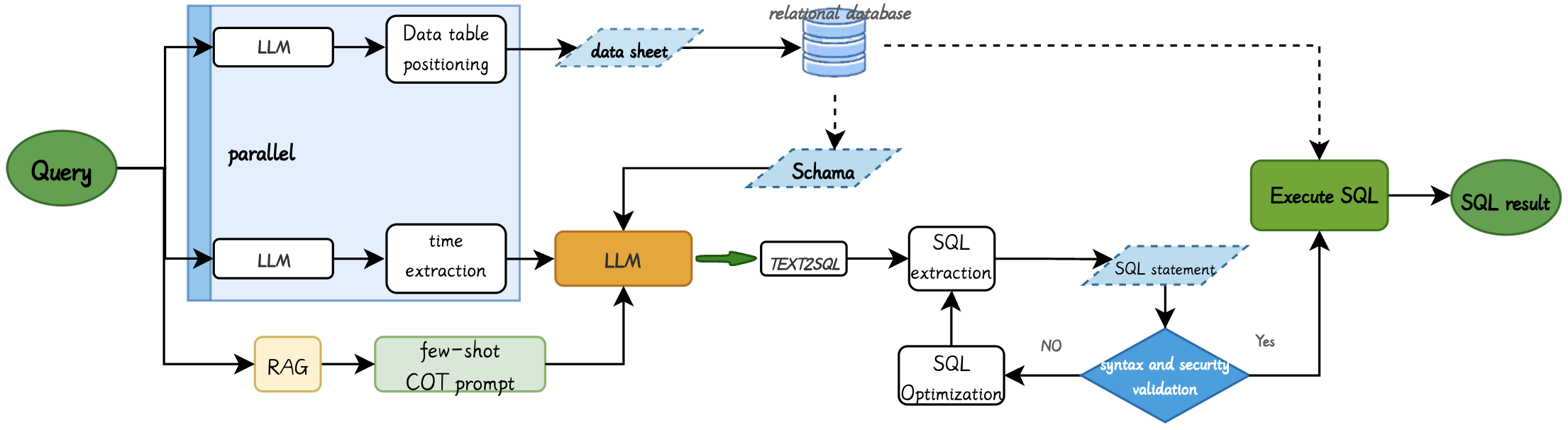}  % 插入图片并设置宽度
    \caption{Text2SQL System Architecture Diagram}  % 图片标题
    \label{fig:en_image/5.png}  % 图片标签，用于引用
\end{figure}
\begin{figure}[ht]  % 强制图片尽量放置在当前页面
    \centering  % 使图片居中
    \includegraphics[width=0.8\textwidth]{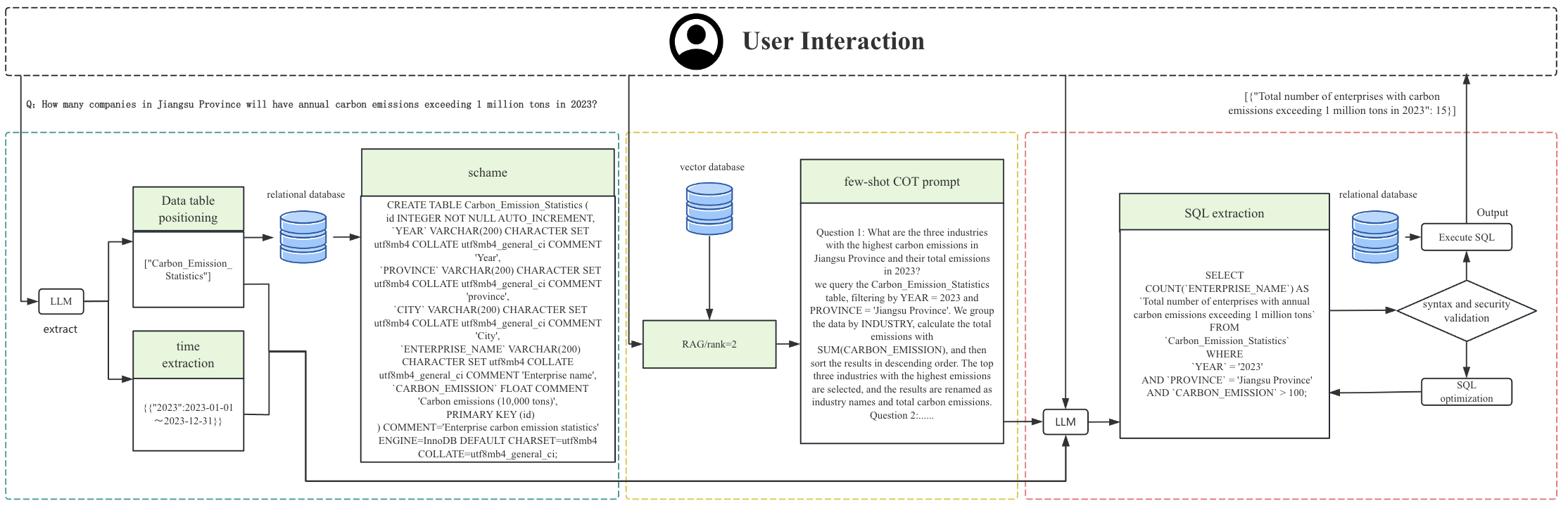}  % 插入图片并设置宽度
    \caption{Text2SQL System Application Example Diagram}  % 图片标题
    \label{fig:7.png}  % 图片标签，用于引用
\end{figure}
\subsection*{Global Climate Change Knowledge Q\&A System}
\subsubsection*{Analysis strategy}
To ensure the scientific, accurate, and practical nature of the enterprise carbon emissions analysis system, the system's analysis dimensions are carefully designed to incorporate international standards from the GHG Protocol, the operational characteristics of the enterprise, data traceability, complexity management, and personalized needs.\par
The analysis dimensions are based on the three emission scopes defined by the GHG Protocol (Scope 1, Scope 2, and Scope 3), comprehensively covering the enterprise's direct emissions (such as emissions from fixed facilities and mobile equipment), indirect emissions (such as the use of purchased electricity), and supply chain emissions (such as upstream and downstream logistics and procurement). By clearly defining emission activities, the system ensures the comprehensiveness and compliance of the accounting process. Additionally, the system incorporates industry-specific design, such as focusing on energy consumption and supply chain activities in manufacturing, and emphasizing employee commuting and waste management in the service industry.\par
The system also analyzes the boundaries of the enterprise's carbon emissions inventory (such as organizational structure, operational boundaries, and business scope), verifying whether all relevant emission sources are included in the report, and reviewing any excluded emission activities and their disclosure reasons. This verification ensures the completeness of the emissions inventory and helps assess whether the enterprise meets the relevance requirements for providing decision-making support to both internal and external users. Moreover, the system evaluates whether the enterprise publicly discloses long-term emission tracking information and whether the summarized reports maintain consistency over time, thereby enhancing the timeliness and logical coherence of the reports.\par
Building on this, the system performs a comprehensive analysis by integrating future greenhouse gas target setting, carbon emission risk identification, and management, providing the enterprise with both carbon emissions assessments and future carbon emissions forecasts and controls. The active involvement of climate experts further improves the accuracy of assessments and the relevance of strategies, making the analysis more tailored to the enterprise's actual situation, and enhancing the system's ability to recognize complex emission data.
This multi-layered, multi-dimensional design not only meets global carbon emission accounting requirements but also flexibly adapts to the specific circumstances of each enterprise, providing high-quality analysis results for decision-makers.\par
\subsubsection*{Hallucination analysis}
The large language model has demonstrated great potential in carbon emission analysis and knowledge question-answering systems, but still faces several key challenges in practical applications, particularly the phenomenon of hallucination. Specifically, hallucination can manifest in the following forms: 1) factual inconsistency, where a certain carbon emission standard may be incorrectly stated as inconsistent with the actual regulations; 2) query inconsistency, where other aspects of carbon emissions may be discussed but the model fails to address the specific question posed by the user; 3) topic deviation, where certain carbon emission-related concepts may be discussed, but the model does not delve into the specific details directly related to the question.
To address these issues, the generated responses of the large model were analyzed and optimized in depth, focusing on improving factual recall, understanding context, and enhancing reasoning ability, thereby ensuring the authenticity, consistency, and verifiability of the generated content. The following work was carried out:\par
\begin{itemize}
\item Constructing document trees and multimodal annotations: Each record in the knowledge base retains the full path and specific location of the source document (e.g., page number, paragraph number). By combining multimodal annotation methods for text, tables, and visual content, the system is able to handle structured data and trace the sources of unstructured text and visual content, as detailed in Diversified Index Module Construction module.\par
\item Self-prompt retrieval architecture: Using a self-prompt retrieval architecture to guide the large model through gradual reasoning, as detailed in Self-Prompting Enhanced Retrieval-Generation Architecture module.\par
\item Timestamp and version control: To handle the potential updates to corporate reports over time, each record in the knowledge base is assigned a timestamp and version information to ensure that users are aware of the timeliness of the data. The system also allows for differential comparisons between different versions of the report, tracking historical content changes.\par
\item Hallucination tagging function: In the process of generating answers, the carbon emission analysis module specifically designed a "hallucination tagging" function. This function verifies whether the generated response is contained in the evidence sentences from the report, screening and tagging any information that deviates from or exceeds the original content to ensure that all responses are based on authentic and verifiable content. 
\end{itemize}
\subsubsection*{Pipeline}
This paper constructs a scientifically robust and highly practical carbon emission report analysis system through the integration of three modules: report summary, report evaluation, and personalized question answering. Specific examples are provided in Appendices 3 and 4. The overall process flow is shown in Figure 1.\par
Report Summary Module:The report summary module aims to assist users in efficiently understanding and analyzing corporate carbon emission reports, particularly focusing on systematically summarizing the disclosure content of companies according to the GHG Protocol guidelines. By integrating multi-source information and extracting relevant report content, along with background data (from the vector database) and company information (from the relational database), the system generates structured and traceable responses. Through summary and key extraction, users can quickly grasp the core information without reading the report word by word, significantly improving report reading efficiency. By using the information from the target report as analytical prompts, this approach effectively reduces hallucinations, ensuring answer accuracy and a clear correspondence with the source data.\par
The report summary module also promotes report compliance. By systematically analyzing the recommendations in the GHG Protocol, the system helps companies identify potential deficiencies in their environmental reports and provides references for improving the content of these reports. It enhances the normative and comprehensive nature of future disclosures, supporting companies in formulating more effective carbon emission strategies. The module achieves end-to-end optimization from data extraction to information summarization through intelligent retrieval, contextual integration, and closed-loop answer generation. The design of the prompt templates ensures user-friendliness while enhancing multi-scenario adaptability, fully leveraging the role of report summaries in decision support. The prompt templates for this module are provided in Appendix 2.The analysis results are shown in Appendix 3.\par
Report Evaluation Module:The GHG Protocol provides detailed disclosure guidelines for each recommendation, specifying the types and granularity of information that companies need to disclose in their reports. To assess whether a corporate carbon emission report complies with the GHG Protocol, a report evaluation module has been designed to analyze the extent to which the report follows the GHG Protocol. For each GHG guideline content, the report evaluation module receives relevant context from the vector retrieval module. It then compares and evaluates this content against the corresponding GHG guidelines, generating an analytical paragraph and a GHG Protocol compliance score ranging from 0 to 100. The prompt template for this module is provided in Appendix 2.The analysis results are shown in Appendix 4.\par
Customized Question Module:This module provides users with the ability to conduct personalized analysis beyond the GHG Protocol framework. It can receive user-defined questions and retrieve relevant contextual information based on the self-prompting retrieval-augmented generation (RAG) architecture to address personalized analysis needs. The module features adaptive prompt templates, specifically optimized for noisy scenarios, reducing redundant information and enhancing semantic parsing capability, thereby ensuring the scientific accuracy and precision of personalized analyses. This flexibility enables the system to better handle complex issues, offering reliable customized analysis support to users. The prompt template for this module is provided in Appendix 2.
\section*{Experiments}
The system uses Qwen-max as the foundational large language model for conducting the experiments and analyses presented in this paper. Langchain is employed to manage Qwen API calls and vector database retrievals. The BGE-reranker-large reranking model is utilized. The text embedding BGE-M3 from Zhiyuan AI is used to embed text blocks, and the Chroma vector database is employed for storing the vectorized text blocks. If the prompt becomes too lengthy (e.g., exceeding 3000 tokens) after inserting retrieved background information, the background information is summarized by progressively removing the least relevant blocks until the prompt fits within the context window. The temperature for all LLM calls is set to 0, and a static vector database is reused for each report.
\subsection*{Data construction}
This paper systematically collects and organizes 1,000 policy and regulatory documents related to corporate carbon emissions management issued by the Chinese government (including both central and local governments) between 2018 and 2024, along with 1180 QA pairs annotated as a test set. Additionally, 100 environmental reports from publicly listed companies are collected from various public resources and databases. These reports typically include information on greenhouse gas emissions, The goal of reducing carbon emissions, implementation measures, and progress in emission reductions during company operations. The data sources include annual and sustainability reports published by companies such as Apple and Google, which provide detailed accounts of greenhouse gas emissions and environmental measures. Another important data source is carbon disclosure platforms, such as CDP (Carbon Disclosure Project) \cite{CDP2023} , which provide extensive carbon emissions data from publicly listed companies. Other sources include government and regulatory publications, such as those from the China Securities Regulatory Commission and the EU’s Sustainable Finance Disclosure Regulation (SFDR) standards.\par
To evaluate the performance of content-aware questions, this paper constructs a test dataset, manually annotating 2,133 high-quality QA pairs based on two real business databases. The dataset covers types such as filtering, aggregation, multi-table joins, and nested queries, emphasizing the context dependence of tabular content. It requires the model to generate accurate SQL queries by combining table structures and instance data. All QA pairs are annotated and reviewed by domain experts to ensure high consistency between questions and answers, providing a reliable evaluation tool for Text2SQL technology.\par
\subsection*{Evaluation indicators}
ROUGE (Recall-Oriented Understudy for Gisting Evaluation) \cite{Lin2004ROUGEAP} is an automated method used to evaluate the quality of generated text, widely applied in tasks such as text summarization, translation, and generation. It primarily measures the quality of text generation by calculating the overlap of vocabulary between the generated text and reference text. ROUGE-N is used to calculate the overlap of N-grams, with the following formula:\par
$$\mathrm{ROUGE-N}=\frac{\sum_{i=1}^N\mathrm{count}(\mathrm{ref}\cap\mathrm{sys})}{\sum_{i=1}^N\mathrm{count}(\mathrm{ref})}\    (3)$$
Here, "ref" represents the reference text, "sys" represents the generated text, "$\cap$" denotes the intersection between the two, and "count" refers to the number of N-grams in the intersection. The core idea of ROUGE-N is to measure the matching ratio of N-grams between the two texts, which reflects the coverage of the reference text by the generated text.\par
Additionally, ROUGE-L is based on the Longest Common Subsequence (LCS) measure, focusing on the matching of word order and considering both semantic and structural similarity. These metrics provide comprehensive automated evaluation tools for text generation tasks.\par
$$\mathrm{ROUGE-L}=\frac{2\times\mathrm{LCS}(\mathrm{ref},\mathrm{sys})}{L_{\mathrm{ref}}+L_{\mathrm{sys}}}\    (4)$$
Here, LCS(ref, sys) represents the length of the Longest Common Subsequence (LCS) between the reference text and the generated text, while Lref and Lsys represent the lengths of the reference and generated texts, respectively.\par
BERTScore is a semantic-based text evaluation method \cite{Zhang2019BERTScoreET} that leverages the pre-trained contextual embeddings of the BERT model to assess the semantic similarity between the generated text and the reference text using cosine similarity. Specifically, BERTScore first generates embedding vectors for both the generated text and the reference text using the BERT model, and then calculates the maximum matching similarity of the embedding vectors for each word in the two texts.
BERTScore mainly includes the following three metrics: \par
Recall (R): It calculates the sum of the maximum similarity of each word in the generated text with the words in the reference text, and then divides it by the total number of words in the reference text. Here, sim(x, s) represents the cosine similarity between the words in the reference text and the generated text, while ref and sys represent the word counts of the reference text and the generated text, respectively.\par
$$\text{Recall (R)}=\frac{\sum_{x\in\text{ref}}\max_{s\in\text{sys}}\ sim(x,s)}{|\text{ref}|}\    (5)$$\par
Precision (P): It calculates the sum of the maximum similarity of each word in the reference text with the words in the generated text, and then divides it by the total number of words in the generated text.\par
$$\text{Precision (P)}=\frac{\sum_{s\in\mathrm{sys}}\max_{x\in\mathrm{ref}}\ sim(s,x)}{|\mathrm{sys}|}\    (6)$$\par
F1 Score: The F1 score is the harmonic mean of precision and recall, which comprehensively considers the balance between the two.\par
$$\mathrm{F1}=\frac{2\cdot\mathrm{P}\cdot\mathrm{R}}{\mathrm{P}+\mathrm{R}}\    (7)$$\par
Execution Accuracy (EX) \cite{Xu2024TCSRSQLTT} compares the execution outputs of the predicted SQL query with the execution outputs of the target SQL query for a given problem. Since there may be multiple valid SQL queries for a single question, the EX metric matches SQL queries that produce the same execution results, disregarding differences in the string representation of the generated SQL and the target SQL.\par
Exact Match (EM) \cite{Yu2018SpiderAL}compares each clause in the predicted SQL query with the corresponding clause in the gold-standard SQL. In SQL, each keyword is treated as a component and decomposed into a set of subcomponents. For example, the SQL query `SELECT avg(col1), max(col2), min(col1)` would be decomposed into the sets `(avg, min, col1)` and `(max, col2)`, and then the sets in the predicted SQL and the actual SQL are compared for consistency. If all components of the predicted SQL match those of the actual SQL, the predicted SQL is considered correct. Although exact match requirements may not always be fully met, some predicted SQL queries may still execute to produce the correct results. Therefore, the EM metric is used as a reference result.\par
\subsection*{Results and analysis}
Through ablation experiments, the paper of the proposes method and its individual modules in improving model performance is further validated. A total of five configurations were compared: standard RAG, the introduction of structured chain-of-thought, diversified indexing, hybrid retrieval, and the final model incorporating all components of our work. The standard RAG, as the baseline method, directly relies on the retrieval-augmented generation framework. Based on the performance of several large models in evaluations on Hugging Face, the following models were selected for experimental comparison: ChatGPT-4o-2024-05-13, Qwen-Max, GLM-4\cite{Zeng2024ChatGLMAF}, Spark 4.0 Ultra \cite{Xinghuo2024}, Baidu ERNIE-4.0-Turbo \cite{BaiduCloud2024}, and Llama-3.1-70B-Instruct \cite{Dubey2024TheL3}.\par
As shown in Table 2, the experimental results indicate that the ROUGE-1, ROUGE-2, and ROUGE-L scores for the standard RAG are 0.529, 0.392, and 0.468, respectively, and the BERTScore Precision, Recall, and F1 scores are 0.826, 0.837, and 0.831. Although the standard RAG exhibits basic answering capabilities, its performance in handling complex questions is more limited. After integrating all the proposes modules, the model performance reaches its optimal level, with ROUGE-1, ROUGE-2, and ROUGE-L scores of 0.592, 0.450, and 0.540, respectively, and BERTScore Precision, Recall, and F1 scores of 0.887, 0.924, and 0.906, which significantly outperform the performance of individual module configurations.\par
In conclusion, the method achieves the best performance under the collaborative effect of all modules. In particular, the structured chain-of-thought module excels in providing logical answers to complex questions, the diversified indexing module effectively optimizes retrieval quality, and the hybrid retrieval module further enhances the system's comprehensive query capability.\par
\begin{table}[h]
\centering
\resizebox{\textwidth}{!}{ 
\begin{tabular}{|c|c|c|c|c|c|c|c|}
\hline
\textbf{Backbone} & \textbf{Methods} & \textbf{ROUGE-l} & \textbf{ROUGE-2} & \textbf{ROUGE-L} & \multicolumn{3}{c|}{\textbf{BERTScore}} \\ \cline{6-8} 
 & & & & & \textbf{Precision} & \textbf{Recall} & \textbf{F1} \\ \hline
\textbf{Spark 4.0 Ultra}\cite{Xinghuo2024} & Standard RAG & 0.527 & 0.370 & 0.454 & 0.808 & 0.825 & 0.817 \\ \cline{2-8} 
 & Structured COT & 0.550 & 0.382 & 0.472 & 0.805 & 0.828 & 0.816 \\ \cline{2-8} 
 & Diversified Indexing & 0.544 & 0.398 & 0.479 & 0.828 & 0.840 & 0.834 \\ \cline{2-8} 
 & Hybrid Retrieval & 0.544 & 0.377 & 0.461 & 0.818 & 0.836 & 0.827 \\ \cline{2-8} 
 & Self-Prompting RAG & 0.576 & 0.406 & 0.492 & 0.820 & 0.881 & 0.849 \\ \hline
\textbf{Llama-3.1-70B-Instruct}\cite{Dubey2024TheL3} & Standard RAG & 0.515 & 0.383 & 0.459 & 0.810 & 0.824 & 0.817 \\ \cline{2-8} 
 & Structured COT & 0.537 & 0.391 & 0.475 & 0.815 & 0.831 & 0.823 \\ \cline{2-8} 
 & Diversified Indexing & 0.541 & 0.399 & 0.478 & 0.830 & 0.841 & 0.836 \\ \cline{2-8} 
 & Hybrid Retrieval & 0.539 & 0.404 & 0.477 & 0.823 & 0.842 & 0.832 \\ \cline{2-8} 
 & Self-Prompting RAG & 0.568 & 0.421 & 0.503 & 0.852 & 0.874 & 0.863 \\ \hline
\textbf{GLM-4}\cite{Zeng2024ChatGLMAF} & Standard RAG & 0.561 & 0.398 & 0.480 & 0.811 & 0.842 & 0.826 \\ \cline{2-8} 
 & Structured COT & 0.572 & 0.405 & 0.485 & 0.799 & 0.850 & 0.824 \\ \cline{2-8} 
 & Diversified Indexing & 0.580 & 0.418 & 0.498 & 0.801 & 0.855 & 0.827 \\ \cline{2-8} 
 & Hybrid Retrieval & 0.588 & 0.412 & 0.491 & 0.803 & 0.840 & 0.821 \\ \cline{2-8} 
 & Self-Prompting RAG & 0.612 & 0.434 & 0.528 & 0.821 & 0.851 & 0.836 \\ \hline
\textbf{ChatGPT-4o-2024-05-13}\cite{Achiam2023GPT4TR} & Standard RAG & 0.521 & 0.388 & 0.463 & 0.812 & 0.826 & 0.819 \\ \cline{2-8} 
 & Structured COT & 0.535 & 0.394 & 0.472 & 0.817 & 0.832 & 0.824 \\ \cline{2-8} 
 & Diversified Indexing & 0.543 & 0.402 & 0.480 & 0.825 & 0.845 & 0.835 \\ \cline{2-8} 
 & Hybrid Retrieval & 0.540 & 0.406 & 0.478 & 0.822 & 0.843 & 0.832 \\ \cline{2-8} 
 & Self-Prompting RAG & 0.570 & 0.425 & 0.506 & 0.855 & 0.877 & 0.866 \\ \hline
\textbf{Baidu ERNIE-4.0-Turbo}\cite{BaiduCloud2024} & Standard RAG & 0.473 & 0.327 & 0.398 & 0.812 & 0.808 & 0.810 \\ \cline{2-8} 
 & Structured COT & 0.527 & 0.387 & 0.465 & 0.831 & 0.837 & 0.834 \\ \cline{2-8} 
 & Diversified Indexing & 0.531 & 0.377 & 0.459 & 0.842 & 0.922 & 0.880 \\ \cline{2-8} 
 & Hybrid Retrieval & 0.533 & 0.393 & 0.474 & 0.832 & 0.839 & 0.835 \\ \cline{2-8} 
 & Self-Prompting RAG & 0.541 & 0.399 & 0.478 & 0.849 & 0.876 & 0.862 \\ \hline
\textbf{Qwen-Max}\cite{Bai2023QwenTR} & Standard RAG & 0.529 & 0.392 & 0.468 & 0.826 & 0.837 & 0.831 \\ \cline{2-8} 
 & Structured COT & 0.552 & 0.409 & 0.493 & 0.849 & 0.924 & 0.885 \\ \cline{2-8} 
 & Diversified Indexing & 0.547 & 0.412 & 0.495 & 0.825 & 0.843 & 0.834 \\ \cline{2-8} 
 & Hybrid Retrieval & 0.541 & 0.399 & 0.478 & 0.830 & 0.841 & 0.836 \\ \cline{2-8} 
 & Self-Prompting RAG & 0.592 & 0.450 & 0.540 & 0.887 & 0.924 & 0.906 \\ \hline
\end{tabular}
}
\caption{\label{tab:model_performance}Self-Prompting Enhanced Retrieval-Augmented Generation Architecture (Text Content) Experiment.}
\end{table}\par
As shown in Table 3, The Text2SQL system achieved an EX score of 89.2\% and an EM score of 79.9\% in the experiments, indicating high accuracy and stability in generating correct SQL queries and matching target SQL queries. Ablation experiments were conducted to analyze the impact of different modules. When the time extraction and table localization modules were removed, the EX score decreased by 21.7\%, and the EM score decreased by 19.6\%. When the few-shot COT prompting was omitted, the EX score dropped by 15.8\%, and the EM score decreased by 17.3\%. In the absence of the SQL optimization module, the EX score dropped by 6.4\%, and the EM score decreased by 7.1\%, highlighting the significant contribution of these three modules to the system's performance.

\begin{table}[ht]
\centering
\begin{tabular}{|l|l|l|}
\hline
Methods & EX(\%) & EM(\%) \\
\hline
Text2SQL System & 89.2 & 79.9 \\
\hline
\hspace*{\tabcolsep}Without time extraction \& table localization & 67.5(↓) & 60.3 \\
\hline
\hspace*{\tabcolsep}Without few-shot COT prompting & 73.4(↓) & 62.6 \\
\hline
\hspace*{\tabcolsep}Text2SQL System & 82.8(↓) & 72.8 \\
\hline
\end{tabular}
\caption{\label{tab:example2}Text2SQL System Experiment.}
\end{table}
\subsection*{Analysis report correlation analysis}

The question-answer relevance refers to the semantic association between the question and the generated answer, evaluated through Large Language Model. In our paper, Qwen-Max is used as the evaluator to assess the semantic correlation between the question and the generated answer. The specific approach involves constructing prompts (see Appendix I), then combining the question and generated answer with the prompt and invoking the large language model to analyze the degree of semantic association between the question and the generated answer, thus determining the accuracy of the generated answer.\par
Additionally, expert scoring is employed to evaluate the responses based on their analysis results, with the understanding of the question being incorporated into the scoring criteria. The experts involved in the evaluation include 10 specialists in carbon emission policy, computer information engineering technology, and energy subfields. We randomly selected 10 reports, each of which was evaluated by 10 experts across 14 analytical dimensions. After statistical analysis of the expert scores, the overall score results are shown in Figure 7. (LLM relevance score (rqa): indicates that the carbon emission analysis system does not use the climate change intelligent question-answering system.)
\begin{figure}[ht]  % 强制图片尽量放置在当前页面
    \centering  % 使图片居中
    \includegraphics[width=0.8\textwidth]{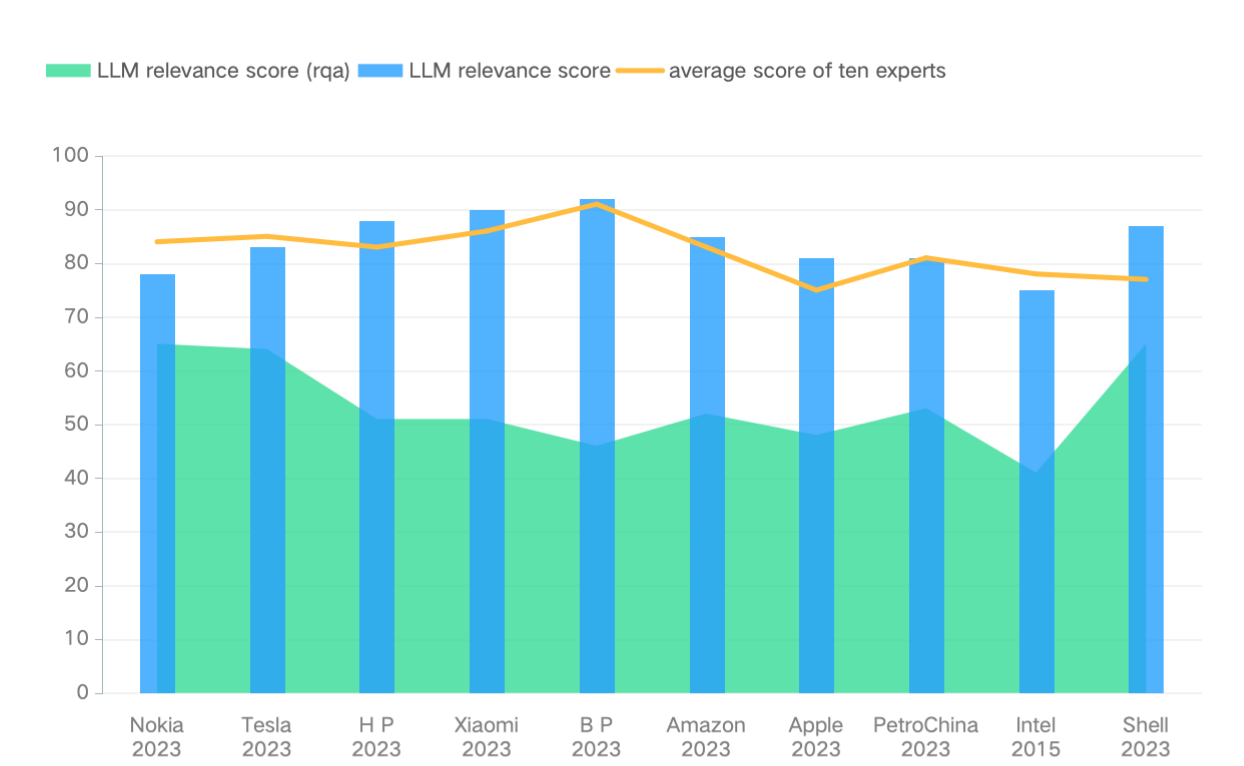}  % 插入图片并设置宽度
    \caption{Text2SQL System Application Example Diagram}  % 图片标题
    \label{en_image/7.png}  % 图片标签，用于引用
\end{figure}
\section*{Conclusions and future work}
This paper proposes a LLM-based enterprise carbon emission analysis and knowledge question-answering system, aiming to enhance the analysis capabilities of carbon emission data and improve the efficiency of knowledge acquisition. The system combines the Self-Prompted Retrieval-Augmented Generation architecture and structured prompt engineering methods, enabling automatic analysis of corporate carbon emission reports and answering knowledge-based questions on global climate change. This helps enterprises provide real-time, accurate information support for policymakers, businesses, and the public, offering comprehensive carbon emission analysis, emission reduction recommendations, and future predictions. It assists businesses in making more scientific and precise decisions in complex carbon emission regulatory environments. Through the intelligent question-answering system, users can easily access climate change-related data and obtain guidance based on the latest research and policies.\par
In future work, the following areas will be further optimized:\par
Localization of LLM Development: Transitioning from the existing Qwen model to a localized large language model to enhance the controllability and reliability of outputs, making it more suitable for carbon emission analysis tasks across various domains. This will involve fine-tuning the model using high-quality datasets evaluated by experts to improve its performance in climate-related tasks.\par
Optimizing Natural Language Processing Capabilities: Further enhancing the natural language processing module, especially in handling complex problems, to reduce hallucination issues that may arise during the parsing and generation process.

\section*{Data and Code availability}
Our code and data will be made public after the paper is accepted.

%Bibliography  % 参考文献
\bibliographystyle{unsrt}  % 参考文献样式
\bibliography{references}  % 引用参考文献文件

\section*{Appendix}
\subsection*{Appendix 1}
P1: You are an expert in the dual-carbon field. Based on the background information, filter out the most relevant rules or content to answer the question, maintaining their integrity and directly quoting the original text. Do not modify it. If no suitable content is available to answer, do not generate anything. The candidate information is as follows: {context} , Question: {query} \\

P2: You are a senior scholar in the dual-carbon field. Your job is to help users better understand the issue without directly answering the question. Please briefly analyze the question following these steps: First, summarize the background information related to the question; second, list the potential key elements or influencing factors; finally, summarize the possible challenges or considerations. Keep it under 100 words and try to include enough background information. Question: {query}\\

P3: Extract key sentences from the following text: {text}. Key sentences should be words or phrases that summarize the main content. Select sentences that reflect the core theme or concept, avoiding common irrelevant words (e.g., conjunctions, adverbs, etc.). The number of key sentences should not exceed 8.\\

P4: You are an expert in the dual-carbon field. Please determine if the question is related to national carbon emission policies and whether it requires querying the enterprise database to answer this question. {query} Output: Does not require / Related to policy / Requires query / Related to policy and requires query\\

P5: Input query: {query} Please follow the output requirements step by step: Output requirements:
1.Break down the query into multiple simple and clear sub-queries.
2.Identify and resolve ambiguities in the query, providing the most likely explanation.
3.Extract the core intent and conceptual elements of the query.
4.Generate a high-level simplified representation, retaining the essence of the query.\\

P6: As a MySQL expert, for the given problem and schema information, your task is to create an accurate and efficient MySQL query to answer the user's question. Generate the SQL that satisfies the requirements based on the given problem and known information. Problem: {query}, Known {time\_info}, use this date directly without calculation. Known Information:Table information: {table\_info}.Generate based on the following information:{remarks}.Reference example: {sample} Requirements: Within the given table scope, query the columns necessary to answer the question. Pay attention to the units of the columns. The SQL must correctly use the column names present in the table. Set Chinese aliases for the column names or calculated results based on the schema and problem. If the query date is not specified but the table has a date column, query the data for the latest date in the table. You can think step by step to generate the correct SQL, but only provide the MySQL statement. Do not include anything else besides the generated SQL.\\

P7: Extract the time period from the question, recording the start and end times. Question: {query}. Known current {now\_date\_info} Return a JSON in the format: {{ Time-related descriptions in the question, accurate to the day: yyyy-MM-dd~yyyy-MM-dd}}. Note: If there are terms like recently, currently, etc., return {{}}. Example: How was the subway operation yesterday afternoon at 4 PM? Result: {{Yesterday: 2024-03-01~2024-03-01}}. Example: How was the subway operation at the beginning of last month? Result: {{Beginning of last month: 2024-03-01~2024-03-10}}. Question: How did this year’s sales compare to last year? Result: {{"This year":2024-01-01~2024-12-31, "Last year":"2023-01-01~2023-12-31"}}. Only output JSON; if no time-related content, return {{}}. Do not explain or output intermediate processes.\\

P8: The current query does not match any carbon emission data. Please check whether your question includes the following elements: company name, year, specific carbon emission type, etc. If there are still issues, please rephrase the question or contact the administrator for assistance.\\

P9: Submit the generated query script into the large model for syntax validation. If errors occur, correct them.\\

P10: You are an outstanding carbon emission relevance assessor, capable of precisely measuring the correlation between analysis dimensions and analysis results, and providing reasonable scores. [Task Description] Your task is to assess the relevance of analysis results according to the four core dimensions of corporate carbon emission analysis (e.g., Scope 1, Scope 2, Scope 3, and customized dimensions). The specific process is as follows: You receive two pieces of text: the first is the analysis dimension, and the second is the analysis result. Analyze the logical relationship between the two and determine whether the analysis result closely revolves around the specified dimension. The score range is from 0 to 100, with 0 indicating completely irrelevant, and 100 indicating completely relevant. [Scoring Criteria] The score is based on the coverage and depth of the analysis result related to the dimension content. If the analysis result fully covers the dimension requirements and provides detailed, specific analysis, a score of 10 is given. If the analysis result deviates from the dimension content or lacks details, the score will be lower. Long and short answers should be fairly scored; accuracy and comprehensiveness matter more than length. Text: {Dimension},{answer}\\

P11: You are a professional document verification assistant. Please verify all information about the company in the given response to ensure that all content can be traced back to its original source, avoiding any unverified speculative content. Below is your task:\par
1.Check "Report Content" and "System Response" for any discrepancies or excess information and mark them as "hallucinations."
2.Provide the corrected response to ensure it completely conforms to the report content's facts and semantics.
3.Output the result in the following format: \par[Report Content]: {report\_content} [SQL Execution Result]: {SQL\_result} (If none, indicate no need to query the enterprise database) [System Response]: {generated\_answer} [Hallucination Mark]: Please mark any parts that do not align with the report content and explain why. [Corrected Response]: Please provide the corrected response to ensure it is accurate and in line with the report content. [Reason Explanation]: Briefly explain why the content was marked as hallucination and provide the correction basis.\\

\subsection*{Appendix 2}

\textbf{Get basic company information prompt:} 
\begin{figure}[H] 
    \centering
    \includegraphics[width=1\textwidth]{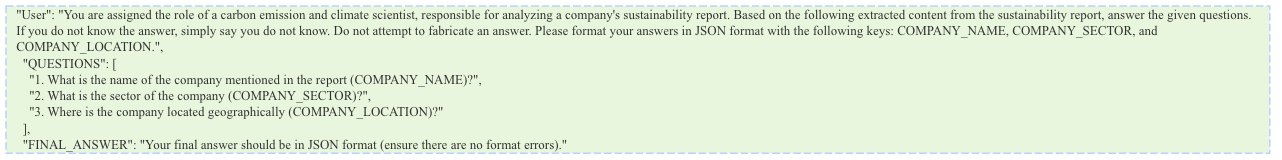}  
\end{figure}

\textbf{Report summary module prompt:} 
\begin{figure}[H] 
    \centering
    \includegraphics[width=1\textwidth]{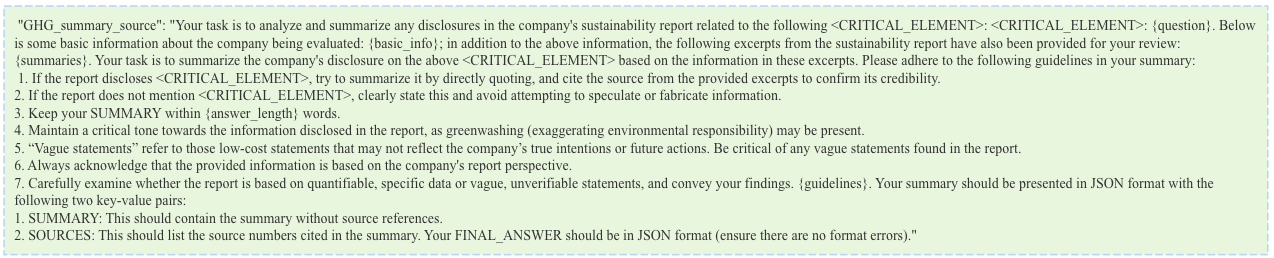}  
\end{figure}
\begin{figure}[H] 
    \centering
    \includegraphics[width=1\textwidth]{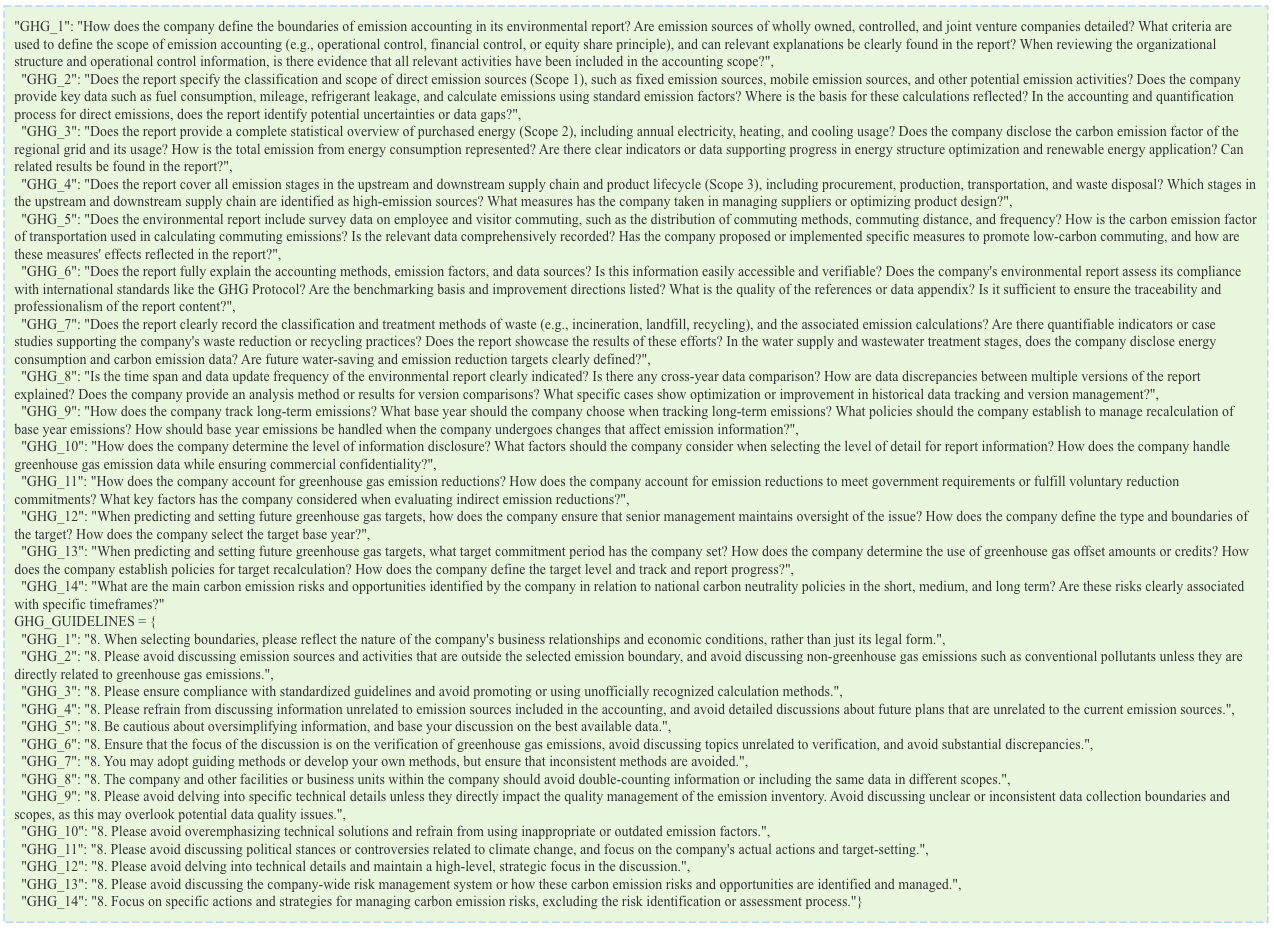}  
\end{figure}
\textbf{Report assessment module prompt:} 
\begin{figure}[H] 
    \centering
    \includegraphics[width=1\textwidth]{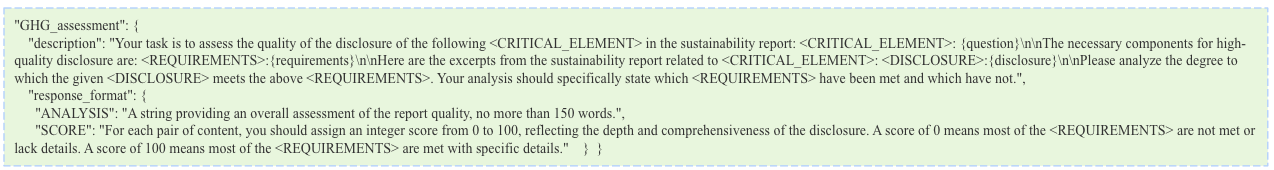}  
\end{figure}
\begin{figure}[H] 
    \centering
    \includegraphics[width=1\textwidth]{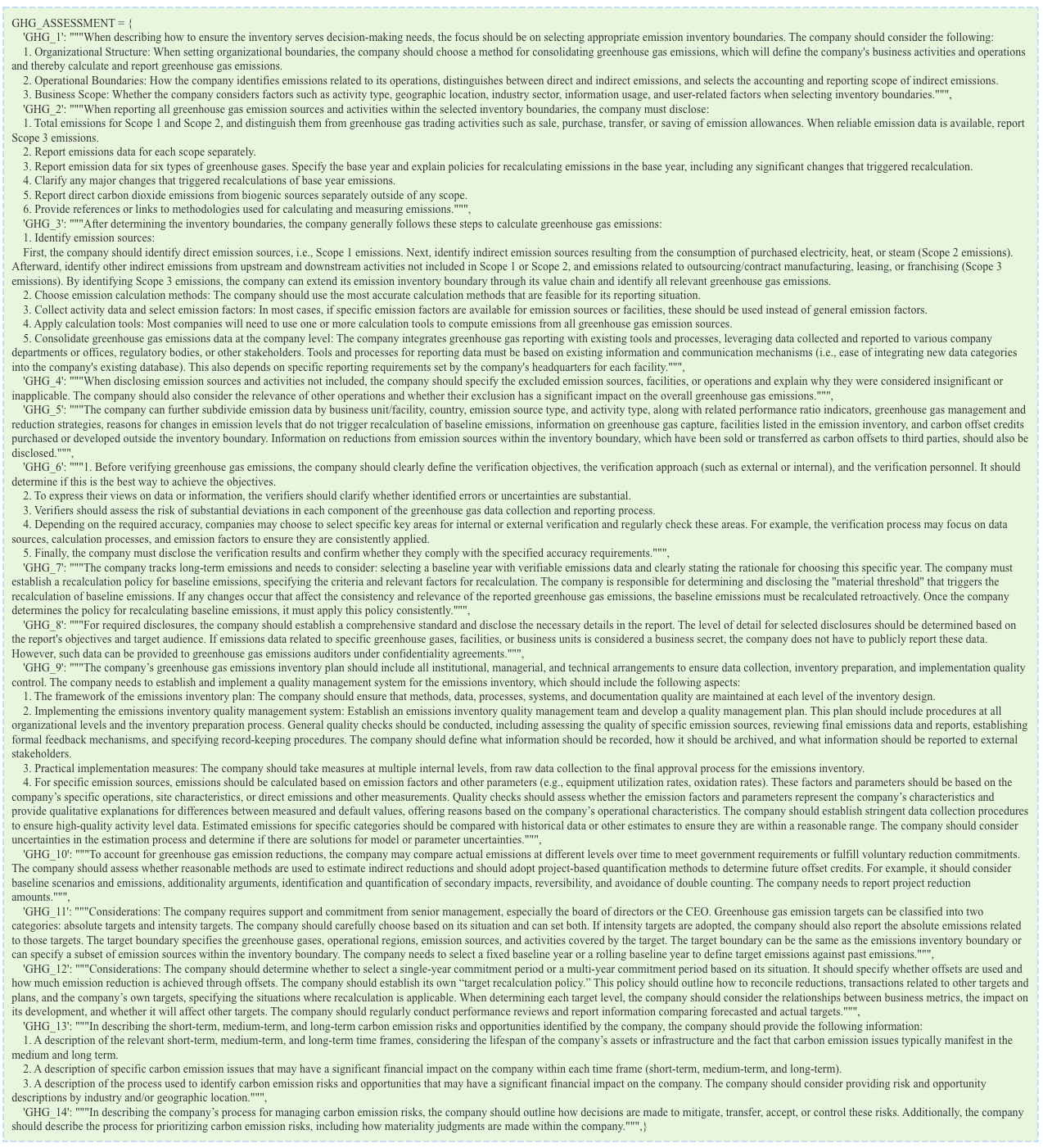}   
\end{figure}
\begin{figure}[H] 
    \centering
    \includegraphics[width=1\textwidth]{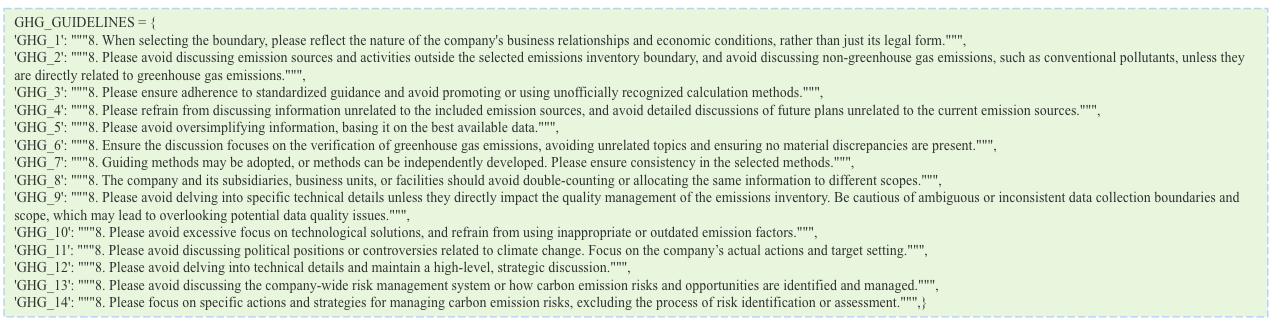}  
\end{figure}
\textbf{Customized question prompt:} 
\begin{figure}[H] 
    \centering
    \includegraphics[width=1\textwidth]{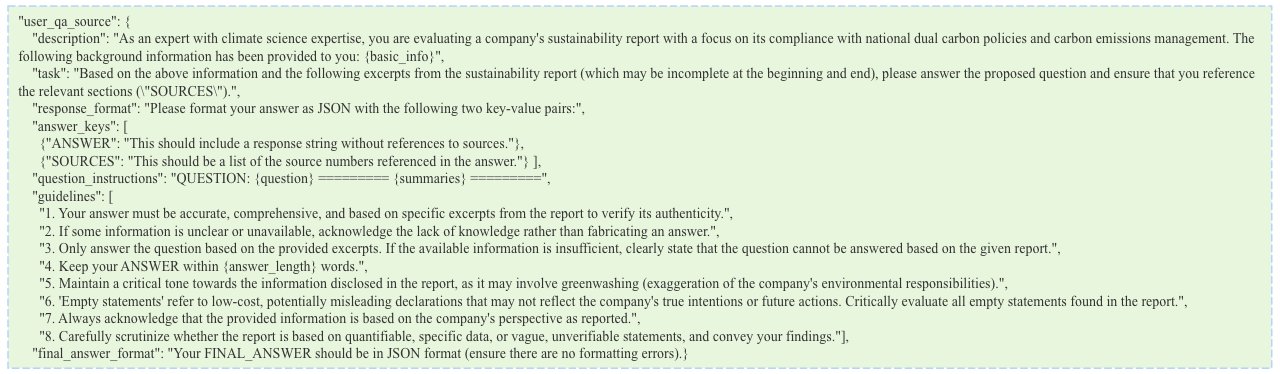}  
\end{figure}
\subsection*{Appendix 3}
\textbf{Take Nokia’s climate and ESG strategy as an example show the results of corporate carbon emissions analysis (answer\_length= 150)} \par
GHG\_1\par
A1: Based on the provided information, the company appears to have defined its emission accounting boundaries, focusing primarily on emissions from its facilities, fleet, and company-owned vessels (Scope 1), market-based emissions from purchased energy (Scope 2), and emissions from the use of sold products by customers, logistics, and emissions from assembly plants and leased vessels in the supply chain (Scope 3). This information is presented in the report as specific descriptions of emission scopes. The report also mentions the emission sources from wholly-owned, controlling, and joint-venture companies, and clarifies the specific role of these companies within its operations and value chain. The delineation of emission accounting boundaries is based on principles of operational control, financial control, or equity share. However, the report does not explicitly mention the specific evidence or detailed information supporting these boundary definitions.

GHG\_2\par
A2: Based on the provided report, Nokia has outlined the classification and scope of its direct emissions sources (Scope 1), including emissions from facilities, fleet, and company-owned vessels. The report mentions key data such as fuel usage and mileage, and emissions have been calculated using standard emission factors. However, the report does not explicitly indicate whether potential uncertainties or data gaps have been identified. Additionally, Nokia discusses collaboration with suppliers to set targets and encourages the use of more sustainable fuels and decarbonized electricity. The report also mentions the use of mixed emission factors, which suggests that the company considered additional factors, rather than relying solely on global average emission factors, when calculating Scope 3 Category 11 emissions. This provides data support for a more accurate reflection of the company’s actual emissions. However, to confirm the final compliance status and potential uncertainties or data gaps, more detailed data and analysis would be required.

GHG\_3 \par
A3: The report partially discloses the use of purchased energy (Scope 2), including the annual consumption of electricity and thermal energy, but does not explicitly mention the use of cooling energy. The report mentions the carbon emission factors for the regional electricity grid and their usage, but specific data is not provided. The total emissions from energy consumption are calculated using a mixed emission factor, which includes a combination of customer-specific emission factors, national average emission factors, and global average emission factors. Progress in optimizing energy structure and examples of renewable energy applications are reflected in the report, such as the company's goal of purchasing 100\% renewable electricity for its facilities. Although the report does not provide all the indicators and data to support these progress statements, the company's efforts to achieve these goals can be inferred from the mentioned plans and actions.

GHG\_4\par
A4: According to the report, Nokia's sustainability report covers emissions from Scope 1, 2, and 3, including procurement, production, transportation, and waste management. Notably, suppliers have committed to reducing their greenhouse gas emissions from manufacturing by 100\% by 2030, which represents a significant management initiative for high-emission sources in the upstream supply chain. Additionally, Nokia collaborates with logistics service providers to encourage the use of more sustainable fuels and works with energy companies to promote decarbonization. Nokia is also focused on reducing the embodied emissions of its products through design and innovation, such as increasing the use of recycled materials and creating circular products. However, the report provides vague descriptions of specific data and measures, lacking detailed quantitative data and specific timelines. This could make it difficult to assess the actual implementation of its commitments.

GHG\_5\par
A5: Based on the provided report, there is no explicit mention of survey data regarding employee and visitor commuting, such as the distribution of commuting modes, commuting distance and frequency, or how carbon emission factors for transportation modes are applied. However, the report highlights the company’s efforts to reduce greenhouse gas emissions from operations and facilities and collaborates with the supply chain to improve energy and resource efficiency. This suggests that the company may have recorded relevant data but has not specifically disclosed it in the report. Additionally, the company mentions setting goals and taking actions to promote climate engagement with suppliers, as well as working with suppliers to set emission reduction targets, but it does not clarify whether specific measures to promote low-carbon commuting have been implemented or how their impact is reflected in the report.

GHG\_6\par
A6: The report partially outlines the accounting methods, emission factors, and data sources, but does not clearly state that these are easily accessible and verifiable. The company mentions emissions from Scope 1, 2, and 3, as well as the use of mixed emission factors. However, the specific accounting boundaries, detailed categorization of emission sources, and the calculation process are not thoroughly explained. Regarding compliance with international standards such as the GHG Protocol, the company refers to its acceptance of the SBTi and some target-setting efforts, but does not explicitly list the benchmarking criteria or areas for improvement. The quality of the references or data appendices appears to be insufficient to ensure the traceability and professionalism of the report's content. 

GHG\_7 \par
A7: The content extracted from the sustainability report covers the company’s records on waste classification, handling methods, and emissions accounting. The company mentions actively managing greenhouse gas emissions from its operations and facilities, as well as collaborating with suppliers to improve energy and resource efficiency across the supply chain. However, the report does not provide specific details on the classification of waste, the handling methods, or the related emissions data. Additionally, there is no clear information on energy consumption and carbon emissions data for water supply and wastewater treatment processes, nor are there specific targets for water conservation and emission reduction in the future.

GHG\_8 \par
A8: The time span and data update frequency are not explicitly stated in the environmental report. The report mentions a comparison between data from 2019 and 2023, but it does not provide detailed tracking of historical data. There is no clear explanation of the data discrepancies between multiple versions of the report or how these differences should be interpreted. Regarding optimization or improvement cases, the report mentions initiatives such as purchasing 100\% renewable electricity, setting climate goals for suppliers, and collaborating with customers on supply chain programs. The company is working towards setting net-zero emission targets and focusing on managing carbon footprints across Scope 1, 2, and 3. However, the report does not specify how it ensures that data from different facilities and business units are not double-counted or omitted in the report.

GHG\_9\par
A9: The company has set clear goals and policies for tracking long-term emissions. First, Nokia should choose the earliest relevant time point with reliable data as the baseline year, such as 1990 or the year following a significant structural change within the company. This helps the company establish a performance baseline, the baseline year emissions, which can be used to compare current emissions. To consistently track long-term emissions, Nokia may need to recalculate baseline year emissions when significant structural changes occur, such as acquisitions, asset divestitures, and mergers. Additionally, Nokia has established policies to manage its carbon footprint across Scope 1, 2, and 3, and is committed to minimizing its carbon footprint through proactive and ongoing management. When changes within the company affect emissions data, Nokia recalculates the annual emissions for the year in question, ensuring consistency with the recalculated baseline year emissions.

GHG\_10\par
A10: Regarding the company's level of information disclosure, Nokia does not explicitly state the method by which it determines the extent of its disclosure. However, based on the level of detail and the information provided in the report, it can be inferred that Nokia considers multiple factors when determining the scope of its disclosures, including the scope of its business, the impact of its technological solutions, collaborations with asset-intensive industries, and the influence within its operations and value chain. Additionally, in handling greenhouse gas emissions data, Nokia appears to have struck a balance between protecting business confidentiality and ensuring transparency. This is evident in its approach of setting targets with suppliers and using mixed emission factors to process data.

GHG\_11 \par
A11:In the report, Nokia has implemented multiple measures to account for greenhouse gas (GHG) emission reductions. First, they established baseline years, such as the baseline year of 2019 and the emissions data from 2023, and compared the actual emission reductions over time to calculate the reduction in emissions. Second, they accounted for the carbon footprint across Scope 1, 2, and 3 emissions and utilized mixed emission factors to calculate customer-specific emissions. To meet government requirements and fulfill voluntary emission reduction commitments, Nokia focuses not only on reducing emissions from its own operations but also collaborates with suppliers, customers, and stakeholders to set targets and take actions. When evaluating indirect emission reductions, key factors include collaboration with suppliers, customer-specific emission factors, and global and national emission standards.

GHG\_12\par
A12: [Hallucination Mark]: The system's response mentioned, "The company ensures senior management's oversight of target setting by conducting internal trials of greenhouse gas emission trading schemes." This information is not referenced in the report content.
[Corrected Answer]: The company has established other short-term, medium-term, and long-term goals to drive specific actions, supporting and accelerating the achievement of the main SBT targets. The current science-based carbon targets cover emissions from Scope 1, 2, and 3. They have also set a target to achieve net-zero emissions across the entire value chain by 2040, with an accelerated 2030 target to reduce emissions from their own operations. Additionally, they collaborate with suppliers and customers to jointly set and achieve emission reduction targets. In 2023, they made progress in areas such as renewable energy in facilities, average product power consumption, and emissions from total assembly suppliers.
[Reason for Revision]: The part about the "internal trials of greenhouse gas emission trading schemes" was not mentioned in the original report content or the provided materials. Therefore, it is identified as hallucinated information. The corrected answer removes this section to ensure full alignment with the report content.

GHG\_13\par
A13: The commitment period for setting greenhouse gas (GHG) targets by the company is determined based on the type of target and whether it is a short-term or long-term goal. Nokia has set a long-term goal for 2040 and plans to gradually achieve this target through continuous actions and goal setting. Regarding the use of GHG offsets or credits, the company must ensure that any purchased or transferred offsets are not simultaneously counted towards the GHG reduction targets of other organizations. When establishing a policy for target double-counting, the company should consider the issue of double-counting reduction efforts across the entire organization and the impact of carbon trading on target performance. As for determining target levels, the company needs to base them on its average levels and the degree of completion of its set goals. Progress tracking and reporting require regular performance checks and comparison with target reporting information.

GHG\_14\par
A14: Due to the lack of detailed report content, it is not possible to accurately identify Nokia's specific carbon emission risks and opportunities related to national dual-carbon policies in the short, medium, and long term. However, based on the provided information, it is clear that Nokia is actively taking measures to reduce its own operational greenhouse gas emissions, including using renewable energy, reducing facility energy consumption, and encouraging suppliers to reduce emissions. In addition, Nokia is committed to minimizing its carbon footprint by managing Scope 1, 2, and 3 emissions and is exploring ways to accelerate the achievement of its net-zero emissions target. These actions likely constitute Nokia's strategy for addressing carbon emission risks and seizing opportunities across different timeframes.
\subsection*{Appendix 4}
\textbf{Take Nokia Nokia’s climate and ESG strategy as an example to show the results of Report Evaluation(answer\_length= 150):} \par
Evaluate\_Analysis1: The disclosure section of the given report mentions the company's emission accounting boundaries, including Scope 1, Scope 2, and Scope 3 emissions, as well as how the company collaborates with suppliers and customers to manage emissions. While the report does not explicitly mention the specific methodologies or factors (such as the nature of activities, geographical location, industry sectors, etc.) considered by the organization when setting its emission accounting boundaries, it is clear from the report that the company is working to define emissions within its operations and value chain and has taken measures to reduce these emissions. The report also references science-based carbon targets related to the SBTi and how the company is collaborating with stakeholders in its supply chain to reduce emissions. However, the report does not provide detailed information on how the emission accounting boundaries are specifically defined, nor does it clearly list the emission sources from wholly-owned, controlling, and joint-venture companies. Therefore, regarding the disclosure quality of the GHG recommendation points, this section of information is insufficient.\par
Score1: 70\par
Evaluate\_Analysis2: The disclosure in the given report mentions emissions from Scope 1 (direct emission sources), including emissions from facilities, fleets, and company-owned vessels, which aligns with the description in the GHG guidelines. However, the report does not specifically outline the classification and scope of direct emission sources, such as fixed emission sources, mobile emission sources, and other potential emission activities. Key data, such as fuel consumption, mileage, and refrigerant leakage, are also not provided. While the report refers to the calculation of global emission factors and collaboration with suppliers and customers, it does not clearly present the specific process for calculating emissions using standard emission factors. Additionally, potential uncertainties or data gaps are not identified. As a result, most of the GHG guidelines are not fully met.\par
Score2: 40\par
Evaluate\_Analysis3:Based on the provided report's disclosure, the content comprehensively covers the key elements required, clearly describing the company's methods for carbon emission accounting across different scopes and the presentation of results. It includes energy consumption disclosures that can be compared to annual usage, particularly by explaining how a broader range of indirect emissions and their impacts are obtained, as well as the overall emissions involving individual customers. The report notably discusses key parameters and the potential for utilizing decarbonization methods in operations, offering more detailed explanations on purchased energy, mix factors, and data records. This showcases the company's optimized considerations in procurement activities, and it is evident that there is a management framework and successful practical outcomes, as well as a clear presentation of cost metrics. Additionally, this case highlights the company’s commitment to reducing and mitigating negative impacts on the environment, and also emphasizes the value and benefits of effective segmentation in different services.\par
Score3: 95\par
Evaluate\_Analysis4: The disclosure in the given report provides a detailed account of the company's efforts in sustainability, including emissions from Scope 1, 2, and 3, as well as measures in supply chain management and product design. It notably highlights collaboration with suppliers, including emission reduction targets set by key assembly suppliers, the use of more sustainable fuels by logistics service providers, and encouraging suppliers to participate in CDP disclosures and set emission reduction targets. These elements reflect the company’s attention and efforts across all emission stages in both upstream and downstream supply chains and product life cycles, offering specific measures and results. Although the report does not directly identify which emission sources are considered high-emission sources and provide detailed descriptions of these sources, overall, the disclosure meets the majority of the GHG guidelines’ requirements.\par
Score4: 85\par
Evaluate\_Analysis5: According to the provided sustainability report, while the company discusses its efforts to reduce its environmental footprint and greenhouse gas emissions, and includes some calculations of emission factors and collaboration within the supply chain, it does not explicitly disclose survey data on employee and visitor commuting. This includes the distribution of commuting methods, commuting distances and frequencies, as well as the specific use of carbon emission factors for transportation modes in accounting for commuting emissions. Additionally, the report does not mention whether the company has fully documented the relevant data, nor does it provide specific details on measures implemented or planned to promote low-carbon commuting, nor how these measures are reflected in the report's outcomes. Therefore, the report falls short in meeting the GHG guidelines to some extent.\par
Score5: 40\par
Evaluate\_Analysis6: In the report's disclosure, the report partially meets the GHG guidelines. Firstly, it mentions emissions from Scope 1, 2, and 3, as well as the details of calculations using mixed emission factors, which aligns with the description of verification objectives and methods. However, the report does not explicitly identify a substantial assessment of errors or uncertainties, nor does it provide evidence of multiple site inspections. Information about the verifiers and the timing of their involvement is also not included. Additionally, while the report references collaboration with the SBTi to promote climate goals, it does not provide detailed information on the verification process or the level of assurance. Therefore, the report's disclosure quality and verification details still require improvement.\par
Score6: 60\par
Evaluate\_Analysis7:In the given report's disclosure, the report partially meets the GHG guidelines' requirements regarding the classification of waste, treatment methods, and emission accounting. For example, it mentions the company's technological solutions, waste recycling, and emission reduction targets. However, specific practices regarding the selection and reporting of verifiable emission data baseline years, recalculation policies, significant disclosure limits, and the retrospective recalculation of baseline emissions are not clearly outlined or presented in the report's disclosure. While collaboration with suppliers on setting emission reduction targets and improving energy efficiency is mentioned, it does not specify how baseline emissions are determined and recalculated, nor how these recalculations affect the consistency and relevance of the reported greenhouse gas emission information. Therefore, there is a lack of transparency and verifiability in relation to key environmental data.\par
Score7: 60\par
Evaluate\_Analysis8: In the provided report's disclosure, the report mentions the time span, data update frequency, and data differences between multiple versions of the report. Although specific time spans or data update frequencies are not explicitly stated, it can be inferred from the text that the company continuously monitors and strives to reduce greenhouse gas emissions in its operations and facilities, setting short-term, mid-term, and long-term goals related to the SBT (Science-Based Targets). The report also includes a comparison of emission data from 2019 to 2023, as well as emission data related to the SBT. Therefore, the continuity of the environmental report and data comparison is well demonstrated. However, the report does not explicitly showcase specific examples of optimization or improvement in historical data tracking, nor does it provide an analysis method or results for comparing specific versions.\par
Score8: 70\par
Evaluate\_Analysis9: The given report's disclosure provides some basic information on how the company tracks emissions, how it handles emissions after changes, and which baseline year is chosen for reporting. While there are no specific details to ensure the accuracy of the data or address technical processes, this represents an initiative and constructive preliminary effort. Therefore, it demonstrates an initial practice and awareness of the greenhouse gas emissions management system, especially in areas where specific responsibilities and measures have been identified, which partially meets the GHG guidelines. However, due to the lack of concrete implementation measures and quality control plans, it does not fully meet the GHG guidelines' requirements.\par
Score9: 60\par
Evaluate\_Analysis10: In the report's disclosure, the company mentions its practices for determining the level of information disclosure, including collaboration with suppliers, setting targets, and working with stakeholders. Additionally, the company discusses considering Scope 1, 2, and 3 emissions when handling greenhouse gas emissions data and mentions its alignment with the SBTi's net-zero emission goals. However, the report does not specifically explain how indirect emission reductions are evaluated, nor does it provide detailed information on the quantification methods used for projects that will later be used for offset credits. Furthermore, the report does not provide specific data on the accounting of greenhouse gas reductions or compare the actual emissions at different time-related levels.\par
Score10:70\par
Evaluate\_Analysis11: In the given report's disclosure, Nokia provides some description of its greenhouse gas emission accounting methods but does not explicitly mention support and commitment from senior management, particularly at the board or CEO level. Regarding target setting, Nokia mentions both absolute and intensity targets, covering emissions from Scope 1, 2, and 3, with a clear baseline year reported. The description of target boundaries is also relatively clear. Additionally, the report addresses the consideration of indirect emission reductions, such as collaborating with suppliers to set targets and encouraging customers to use renewable energy. However, the report does not provide detailed information on how the target baseline year is selected or how indirect emission reductions are specifically calculated.\par
Score11:75\par
Evaluate\_Analysis12: In the given report's disclosure, the company provides a detailed explanation of its science-based carbon targets (SBT) and the various actions and commitments taken to ensure the achievement of these targets. These actions and commitments include reducing greenhouse gas emissions from facilities, improving energy efficiency, collaborating with suppliers on sustainability, and setting net-zero emissions targets across the entire value chain. Additionally, the company discloses its progress toward these targets at different time points, including short-term and mid-term goals, and specifically outlines its achievements in climate and energy usage. Although the report does not specify whether the company has chosen a single-year or multi-year commitment period based on its circumstances, or whether offset credits are used in setting the targets, it has provided a relatively comprehensive disclosure on the main aspects.\par
Score12:80\par
Evaluate\_Analysis13: In the given report's disclosure, Nokia provides a detailed description of its short-term, mid-term, and long-term greenhouse gas emission targets, including specific action plans, SBTi submissions, and collaboration strategies with suppliers and customers. However, the report does not explicitly provide details on the duration of the commitment period, the use of greenhouse gas offset credits or allowances, the policies for recalculating targets, or how progress is specifically tracked and reported. Additionally, the description of how the company identifies and addresses carbon emission risks and opportunities over different time horizons is not sufficiently detailed.\par
Score13:60\par
Evaluate\_Analysis14:In the report's disclosure, the company mentions its climate-related targets and action plans across different time horizons (short-term, mid-term, and long-term), such as the setting of SBT targets, emissions from Scope 1, 2, and 3, and the actions and partnerships the company has taken to reduce carbon emissions. However, the report does not clearly describe the specific carbon emission risks and opportunities identified by the company, nor does it explain whether these risks are linked to specific time horizons. Additionally, the process for managing carbon emission risks, prioritizing these risks, and making materiality judgments within the company is not sufficiently detailed.\par
Score14:65
\subsection*{Appendix 5}
\textbf{Comparison of different methods to build index modules:} \par
Question: According to the "National Carbon Market Development Report," how does China's national carbon market demonstrate the flexibility and applicability advantages of the carbon market mechanism through its unique allocation method, thus contributing to global carbon market mechanism innovation and attracting significant international attention?\par
method: Qwen-Max\par
Knowledge base recall list:None\par
Answer:The national carbon market in China demonstrates the flexibility and applicability advantages of the carbon market mechanism through its unique quota allocation method, mainly reflected in the following aspects: 
1. Combination of Historical Intensity Method and Benchmark Method: It takes into account both the company’s past emissions intensity and the best practices in the industry, ensuring fairness and incentives.
2. Differentiated Management: It implements differentiated quota allocation strategies based on the characteristics and emission reduction potential of different industries, promoting the optimization of industrial structure.
3. Dynamic Adjustment Mechanism: It adjusts the total quota and allocation rules in a timely manner based on technological advancements and socio-economic development, maintaining the long-term effectiveness of the market.\par

method: RAG\par
Knowledge base recall list:['Shanghai", do a good job in experience sharing and mutual learning, carry out carbon market capacity building activities,... promote international cooperation, and continuously optimize system design.International attention to the national carbon market continues to increase,... innovating the global carbon market mechanism Contributing to the "China Plan". ', 'The national carbon market accounts for more than 40\% of the country's carbon dioxide emissions... The two markets, the mandatory and the voluntary, constitute the national carbon market system. The national carbon market has an important impact on the global carbon price level and trading mechanism. With important influence,… Contribute "China's solution" to the innovation of the global carbon market mechanism ', 'The national carbon market controls greenhouse gas emissions through market mechanisms... helps the economy and society transition to green and low-carbon']\par
Answer:China's quota allocation method based on carbon emission intensity control targets can not only consolidate the responsibility for greenhouse gas emission control on enterprises, but also provide incentives for carbon emission reduction. It demonstrates the flexibility and applicability advantages of the carbon market mechanism, and provides a good foundation for the global carbon market. Mechanism innovation contributes to the “China solution”. China's carbon market has an important influence on the global carbon price level and the effectiveness of the carbon trading mechanism. The construction and operation of the national carbon market have attracted great attention from the international community.\par

method: Self-Prompting RAG\par
Knowledge base recall list:['Document name: The Ministry of Industry and Information Technology issued the "Guidelines for the Preparation of Standards for Carbon Footprint Accounting Rules for Key Industrial Products".pdf, page number: 7On the 30th, the technical review should in principle reach an agreement... Social groups are encouraged to self-declare and disclose on the standard information public service platform . ', 'File name: Interim Regulations on the Management of Carbon Emissions Trading (Order No. 775 of the State Council of the People's Republic of China).otl.pdf, Page number: 13 The ecological environment authorities and other departments with supervisory and management responsibilities,… The person being inspected shall truthfully report the situation and provide information. ', 'File name: National Carbon Market Development Report (2024) .pdf, page number: 6 The national voluntary greenhouse gas emission reduction trading market was officially launched... The national carbon market has contributed the "China Solution" to the innovation of the global carbon market mechanism ". ',']\par
Answer:China's quota allocation method based on carbon emission intensity control targets demonstrates the flexibility and applicability advantages of the carbon market mechanism, and contributes the "Chinese solution" to the innovation of the global carbon market mechanism. China's carbon market has an important influence on the global carbon price level and the effectiveness of the carbon trading mechanism. The construction and operation of the national carbon market have attracted great attention from the international community. China's quota allocation method based on carbon emission intensity control targets can not only consolidate the responsibility for greenhouse gas emission control on enterprises, but also provide incentives for carbon emission reduction. It demonstrates the flexibility and applicability advantages of the carbon market mechanism, and provides a good foundation for the global carbon market. Mechanism innovation contributes to the “China solution”.\par

\section*{Acknowledgements}
This work was supported by Guiding central and local science and technology development funding projects (246Z 0107G), and Shijiazhuang City Science and Technology Plan Key Project (221130321A).

\section*{Additional information}
\textbf{Correspondence} and requests for materials should be addressed to M.H.\\

\end{document}